%% file: sample-sigconf.tex
\documentclass[sigconf]{acmart}
\acmConference{}{}{} 
\acmBooktitle{} 
\acmPrice{} 
\acmISBN{} 
\acmDOI{}

\usepackage{array}
\usepackage{multirow}
\usepackage{url}
\usepackage{xcolor}
\usepackage{xspace}
\usepackage{adjustbox}
\usepackage{algorithm}
\usepackage{algpseudocode}
\usepackage{tikz}
\usepackage{stfloats}
\usepackage{hyperref}

\definecolor{darkpink}{RGB}{255,128,14}
\definecolor{lightblue}{RGB}{0,107,164}
\definecolor{darkgreen}{RGB}{0,158,115}

\newcommand{\memit}{\texttt{MT}\xspace}
\newcommand{\aedit}{\texttt{AE}\xspace}
\newcommand{\zsre}{\textnormal{\texttt{zsRE}}\xspace}
\newcommand{\cf}{\textnormal{\texttt{CF}}\xspace}
\newcommand*\circled[1]{\tikz[baseline=(char.base)]{
    \node[shape=circle,draw,inner sep=1pt] (char) {\small #1};}}

\AtBeginDocument{%
  }

\copyrightyear{2026}
\acmYear{2026}
\setcopyright{cc}
\setcctype{by}
\acmConference[KDD '26]{Proceedings of the 32nd ACM SIGKDD Conference on Knowledge Discovery and Data Mining V.2}{August 09--13, 2026}{Jeju Island, Republic of Korea}
\acmBooktitle{Proceedings of the 32nd ACM SIGKDD Conference on Knowledge Discovery and Data Mining V.2 (KDD '26), August 09--13, 2026, Jeju Island, Republic of Korea}
\acmDOI{10.1145/3770855.3817879}
\acmISBN{979-8-4007-2259-2/2026/08}

\begin{document}

\title{Can Fine-Tuning Erase Edits? On the Fragile Coexistence of Knowledge Editing and Fine-tuning}

\author{Yinjie Cheng}
\affiliation{%
  \department{School of Computer Science}
  \institution{University of Sheffield}
  \city{Sheffield}
  \country{UK}}
\email{ycheng80@sheffield.ac.uk}

\author{Paul Youssef}
\affiliation{%
  \department{Mathematics and Computer Science}
  \institution{Marburg University}
  \city{Marburg}
  \country{Germany}
}
\email{paul.youssef@uni-marburg.de}

\author{Christin Seifert}
\affiliation{%
  \department{Mathematics and Computer Science}
  \institution{Marburg University}
  \city{Marburg}
  \country{Germany}
}
\email{christin.seifert@uni-marburg.de}

\author{Jörg Schlötterer}
\affiliation{%
  \department{Mathematics and Computer Science}
  \institution{Marburg University}
  \city{Marburg}
  \country{Germany}
}
\email{joerg.schloetterer@uni-marburg.de}

\author{Zhixue Zhao}
\affiliation{%
  \department{School of Computer Science}
  \institution{University of Sheffield}
  \city{Sheffield}
  \country{UK}}
\email{zhixue.zhao@sheffield.ac.uk}

\renewcommand{\shortauthors}{Yinjie Cheng et al.}

\begin{abstract}
Knowledge editing (KE) offers a lightweight alternative to retraining for updating large language models (LLMs). Meanwhile, fine-tuning remains the default operation for adapting LLMs to new domains and tasks. Despite their widespread adoption, these two post-training interventions have been studied in isolation, leaving open a crucial question: if we fine-tune an edited model, do the edits survive? This question is motivated by practical objectives: removing covert or malicious edits, and preserving beneficial edits. If fine-tuning impairs edits (Fig.~\ref{fig:teaser}), current KE methods become less efficient, as a newly fine-tuned model requires re-editing; if edits persist, fine-tuned models risk propagating hidden malicious edits, raising serious safety concerns. To this end, we systematically quantify edit decay after fine-tuning across 254 experimental configurations. Our results show that in general, edits decay substantially after subsequent fine-tuning. AlphaEdit exhibits the greatest decay on the zsRE benchmark when applied to GPT-J, where $25.27\%$ of previously successful edits become unsuccessful after fine-tuning. We further find that fine-tuning only the edited layers is sufficient to effectively remove edits, while incurring only modest degradation in downstream performance. Surprisingly, fine-tuning non-edited layers leads to greater edit decay than all-layer fine-tuning. Besides, our activation space analysis reveals that fine-tuning produces a larger and more coherent representational shift, both in magnitude and direction, than KE. Overall, our study underscores the necessity of evaluating KE within the broader LLM application pipeline. Our code and results are available at \texttt{https://github.com/Cheng-Yinjie/edit\_decay\_KDD}.

\end{abstract}




\maketitle

\newcommand\kddavailabilityurl{https://doi.org/10.5281/zenodo.20497129}
\ifdefempty{\kddavailabilityurl}{}{
\begingroup\small\noindent\raggedright\textbf{Resource Availability:}\\
The source code and data of this paper has been made publicly available at \url{\kddavailabilityurl}.
\endgroup
}

\section{Introduction}

\begin{figure}
\centering
\includegraphics[width=0.47\textwidth]{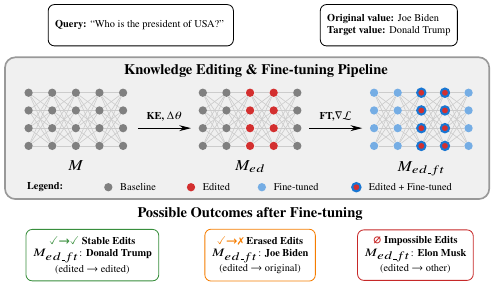}
\caption{Consequences of an LLM ($M$) that undergoes an edit ($M_{ed}$ ) and then fine-tuning ($M_{ed\_ft}$). Edits are classified as \textit{Stable Edits}, \textit{Erased Edits} and \textit{Impossible Edits}. Here is an illustrative example, we show more patterns and real cases in Sec.~\ref{sec:qualitative_analysis}.}
\label{fig:teaser}
\end{figure}

Large Language Models (LLMs) can be updated after pre-training via two main approaches. The first is fine-tuning (FT), where model parameters are updated by training the model on a task-specific dataset~\citep{howard-ruder-2018-universal}. FT also includes parameter-efficient variants such as LoRA~\citep{hu2021loralowrankadaptationlarge} and DoRA~\citep{liu2024doraweightdecomposedlowrankadaptation}, collectively referred to as Parameter-Efficient Fine-Tuning (PEFT). The second approach is knowledge editing (KE)~\citep{mazzia2024surveyknowledgeeditingneural}. Unlike FT, which adapts a model to specific tasks, KE is used to update the model's factual knowledge with limited data and compute budget. Knowledge updated through KE is crucial for keeping the model current and accurate in question‑answering tasks, and its robustness therefore warrants explicit attention, especially when contrasted with model's pretrained knowledge~\citep{meng2023locatingeditingfactualassociations, meng2023masseditingmemorytransformer}. Despite the active research on KE~\citep{wang2023surveyke, mazzia2024surveyknowledgeeditingneural, fang2025alphaeditnullspaceconstrainedknowledge}, and the fact that FT is the de facto approach for adapting LLMs to downstream tasks~\citep{parthasarathy2024ultimate}, no prior work has examined how KE is affected by FT.

This gap has significant implications for real-world model deployment, where LLMs are typically produced through multiple sequential adaptation stages involving heterogeneous techniques. If FT impairs knowledge edits, then more robust KE techniques need to be developed to avoid updating knowledge in every fine-tuned model; conversely, if edits persist, then fine-tuned models may inherit and propagate malicious edits from the base model. This risk is especially concerning given recent evidence that KE can be weaponized for biasing, backdooring~\citep{li2024badedit, chen2024can, youssef2025positioneditinglargelanguage}, or spreading misinformation~\citep{ju2024floodingspreadmanipulatedknowledge} in LLMs. Such ``inheritance'' compromises LLM safety, necessitating robust inspection tools to detect and neutralize adversarial edits.

To address the gap, we examine three representative KE methods (MEMIT, AlphaEdit and MEND) across three FT approaches (full-parameter FT, LoRA, and DoRA) with three KE datasets (see Sec.~\ref{sec:preliminary_setup}). Our findings show that \textbf{FT generally impairs edits}, though larger models exhibit greater structural robustness. We also show that \textbf{FT either only edited or only the non-edited layers both helps remove edits}. Overall, our empirical findings show that existing KE methods fail to produce consistently stable edits that persist under subsequent FT, highlighting the need for KE approaches that are explicitly compatible with downstream adaptation. 

In summary, our key contributions are as follows: 
\begin{itemize}
    \item \textbf{Comprehenssive empirical study of FT on KE:} We conduct the first systematic study of FT’s impact on KE across methods, architectures, and datasets, covering 254 configurations. We introduce an evaluation framework and the \textit{Edit Flip Ratio (EFR)} metric to quantify KE degradation more precisely, serving as a reference for future work.
    \item \textbf{Multi-level mechanistic analysis:} We analyze FT's affect at edit and activation level. At the edit level, we identify four cases (\textit{Stable, Erased, Emergent, Impossible}) and analysis them with knowledge domain and frequency. At the activation level, we show that FT induces larger activation shifts than KE, explaining the fact that FT erases edits.
    \item \textbf{Practical strategies for edit removal:} We show that fine-tuning only edited or non-edited layers removes edits more effectively than tuning all layers, with slight FT performance loss, offering feasible methods for edit-removal.
\end{itemize}

\section{Related work}\label{related_work}

\paragraph{Knowledge Editing.} KE aims to update factual knowledge in LLMs~\citep{youssef-etal-2023-give, petroni-etal-2019-language} without full retraining. Early causal-intervention and direct-weight methods showed that factual associations can be localized and modified~\citep{meng2023locatingeditingfactualassociations,mitchell2022fast}. Scalable multi-edit approaches followed, notably MEMIT \citep{meng2023masseditingmemorytransformer}, which supports thousands of edits, and AlphaEdit \citep{fang2025alphaeditnullspaceconstrainedknowledge}, which constrains perturbations to null spaces to minimize interference with unrelated knowledge. Unlike the above locate-then-edit methods, hypernetwork-based approaches, such as MEND~\citep{mitchell2022fast}, use compact auxiliary editing networks that leverage a single input-output pair to quickly apply localized adjustments to a pre-trained model’s weights~\citep{li2025reinforced}. Parameter-preserving KEs~\citep{zheng-etal-2023-edit, youssef-etal-2026-persuasion} leverage the strong in-context abilities of LLMs~\citep{youssef-etal-2024-queen} or external memories~\citep{mitchell2022memorybasedmodeleditingscale, wang2024wise}. More comprehensive surveys have since consolidated various methods, benchmarks, and evaluation challenges, providing a broad overview of the field~\citep{mazzia2024surveyknowledgeeditingneural, wang2023surveyke}. 

Evaluation has been focused around two datasets: \cf and \zsre, using metrics that assess direct editing success (efficacy), paraphrase generalization (efficacy with paraphrased queries), and impact on non-target knowledge (locality, also known as neighborhood success) \citep{meng2023masseditingmemorytransformer,mitchell2022fast}. Concurrently, several studies have investigated limitations, including instability under sequential/multi-point edits, scope miscalibration, and side effects on unrelated knowledge \citep{mitchell2022memorybasedmodeleditingscale,li2024pitfalls}. KE is also closely connected to unlearning and ``knowledge washing'', which removes or suppresses stored knowledge at scale \citep{wang2025large}. Despite rapid progress, \textbf{most KE studies evaluate edits in isolation}, leaving open whether edits persist when models are subsequently adapted to downstream tasks.

\paragraph{Full and Parameter-Efficient Fine-tuning.} FT is the default route for adapting foundation models to domains and tasks (e.g., ULMFiT; \citealp{howard-ruder-2018-universal}). PEFT techniques have become the practical workhorse across the LLM production pipeline: LoRA injects low-rank adapters into frozen backbones \citep{hu2021loralowrankadaptationlarge}, DoRA decomposes weights into magnitude and direction to better match full FT capacity \citep{liu2024doraweightdecomposedlowrankadaptation}, and adapter families provide modular, swappable components for rapid specialization \citep{hu-etal-2023-llm}. Empirically, PEFT often outperforms few-shot ICL while being dramatically cheaper than full FT \citep{liu2022few}. As a result, PEFT has become the de facto standard across the production pipeline of LLM-based applications. In practice, cloud providers (e.g., Azure and Google Cloud) and model hubs (e.g. HuggingFace) distribute LoRA/adapter checkpoints or supporting pipeline as compact add-ons, enabling organizations to maintain a single backbone and compose task- or client-specific adapters at deploy time \citep{hu2021loralowrankadaptationlarge, ye2023mlora, liu2024doraweightdecomposedlowrankadaptation}. \textbf{This widespread industrial adoption makes understanding PEFT’s interaction with KE highly consequential.}

Despite the active research in and strong performance of FT and KE methods, these two families of techniques have been studied almost entirely in isolation. One line of work has shown that FT can overwrite or ``wash out'' factual associations \citep{wang2025large}, while another has examined whether edits introduced via prompting persist under distributional shifts \citep{zheng-etal-2023-edit}. However, there has been no systematic study of how full FT or PEFT affects the stability of explicitly introduced knowledge edits. This gap is crucial because \textbf{real-world deployment rarely involves static models: models are regularly fine-tuned to new domains and tasks after initial training.}

\paragraph{Malicious Editing.} \label{related_work_safety} Beyond utility, model editing raises important safety concerns. Recent work has demonstrated that editing can be exploited as an attack vector, e.g., by backdooring models through malicious edits \citep{li2024badedit}, injecting harmful content \citep{chen2024can}, or enabling misinformation to spread across multi-agent systems \citep{ju2024floodingspreadmanipulatedknowledge}. Recent work~\citep{youssef2025positioneditinglargelanguage, youssef-etal-2025-fact, youssef-etal-2025-make, 10.1145/3711896.3737001, youssef2026tracing, holmov2026maskruleallhidden} emphasizes the broader safety risks of covert edits persisting through the lifecycle of model adaptation and deployment. If fine-tuned models inherit edited behavior from the original model, then harmful or biased content could silently propagate across production models; if FT impairs beneficial corrective edits, operators may need costly re-editing after every adaptation step. This tension motivates our empirical focus on whether, when, and how edits decay after subsequent FT.

\section{Experiments}\label{sec:experiment}
To this end, we construct four groups of models:  
\circled{1} \textbf{base} models, $M$, no FT nor KE.
\circled{2} \textbf{FT-only} models, $M_{\text{ft}}$; 
\circled{3} \textbf{KE-only} models, $M_{\text{ed}}$;  
\circled{4} \textbf{KE-then-FT} models, $M_{\text{ed\_ft}}$. We compare the editing performance gap between $M_{\text{ed}}$ and $M_{\text{ed\_ft}}$ to assess the impact of FT on KE (Sec.~\ref{sec:results_ed_ft}). We evaluate the downstream performance of $M_{\text{ft}}$ to validate the FT performance in general (Sec.~\ref{sec:reasoning_ability_check}).

Our experiments cover five models, two KE datasets, three KE methods, four editing number settings, and four fine-tuning settings, resulting in 254 independent model configurations. As illustrated in Tab.~\ref{tab:total_experiment_groups}, each column represents a selectable parameter and the full set of experiment configurations $S$ can be obtained through their Cartesian product. 

\input{tabs/app/total_experiment_groups}

\subsection{Models and datasets}\label{sec:preliminary_setup}

\paragraph{Models.}\label{sec:models}
Consistent with prior work on KE~\citep{meng2023locatingeditingfactualassociations, fang2025alphaeditnullspaceconstrainedknowledge}, we include GPT-J-6B~\cite{gpt-j}, GPT2-XL~\citep{radford2019language}, Llama2-7B-hf ~\cite{touvron2023llama2openfoundation}, and Llama3.1-8B-Instruct~\cite{grattafiori2024llama3herdmodels}). Detailed model configurations are provided in our repository. We also evaluated DeepSeek but ultimately excluded it due to its poor KE performance (see App.~\ref{app:ke_about_deepseek}).

\paragraph{KE datasets.}\label{sec:editing_datasets}
Following \citet{meng2023locatingeditingfactualassociations}'s work, we employ CounterFact (\cf)~\citep{meng2023locatingeditingfactualassociations} and Zero-Shot Relation Extraction (\zsre)~\citep{levy-etal-2017-zero, mitchell2022memorybasedmodeleditingscale}. \cf comprises factual and counterfactual statements, each paired with multiple paraphrased prompts; \zsre is a question-answering dataset.

\paragraph{FT datasets.}\label{sec:fine-tune_datasets}
Adhering to established benchmarks for DoRA \citep{liu2024doraweightdecomposedlowrankadaptation, hu-etal-2023-llm}, we utilize the \textit{Commonsense Reasoning dataset}, which comprises seven multiple‑choice question datasets and one yes‑or‑no question dataset. To assess model performance on knowledge-intensive tasks, we additionally include \textit{HotpotQA}, a multi-hop question-answering dataset containing 97, 900 entities.

\paragraph{KE methods.}\label{sec:editing_methods}
We focus on three popular parameter-modifying methods: MEMIT (\memit)~\citep{meng2023masseditingmemorytransformer}, AlphaEdit (\aedit)~\citep{fang2025alphaeditnullspaceconstrainedknowledge} and MEND~\citep{mitchell2022fast}.

\paragraph{FT methods.}\label{sec:fine-tune_methods}
We use LoRA, DoRA, and full-parameter FT with the identical setup used by \citet{hu2021loralowrankadaptationlarge} and \citet{liu2024doraweightdecomposedlowrankadaptation}. 

\paragraph{Evaluation.} \label{sec:eval_methods} 
For KE, we use \textit{Efficacy Success} (ES), \textit{Paraphrase Success} (PS) and \textit{Neighborhood Success} (NS), as described in Sec~\ref{related_work}, with detailed definitions provided in App.~\ref{app:KE_metrics_intro}.

Moreover, we introduce \textit{Edit Flip Ratio (EFR)} as a metric to quantify, at the individual edit level, the number of successful edits that become unsuccessful after fine-tuning. Unlike the ES metric, which measures the overall performance, EFR exclusively focuses on the stability of succeeded edits.

We use a binary indicator $s^\text{M}_i\in\{0, 1\}$ to represent the editing outcome for fact $i$ in model $M$, where 1 indicates the editing is successful and 0 otherwise (e.g., $s^\text{ed\_ft}_i=1$ signifies the $i^\text{th}$ edit remains successful after fine-tuning in model $M_\text{ed\_ft}$). The evaluation criteria for the indicator are consistent with KE metrics (Equ.~\ref{eq:efficacy_def_zsre}-\ref{eq:efficacy_def_mcf} in App.~\ref{app:KE_metrics_intro}). We then define a \textit{flipped case} as an edit that is successful after KE ($s^{\mathrm{M_\text{ed}}}_i=1$), but becomes unsuccessful after fine-tuning ($s^{\mathrm{M_\text{ed\_ft}}}_i=0$). Accordingly, EFR is defined as the number of flip cases in $M_\text{ed\_ft}$ divided by the total number of successful cases before fine‑tuning, as shown in Equ.~\ref{eq:efr_formula_2}.

{
\small 
\begin{equation}
\mathrm{EFR} =  \frac{|\{ s_i | s_i^{\mathrm{M_\text{ed\_ft}}} = 0 \wedge s_i^{\mathrm{M_\text{ed}}} = 1 \} |}{|\{ s_i\} |} 
\label{eq:efr_formula_2} 
\end{equation} 
}

For fine-tuning, we assess the model's reasoning capability using accuracy for both FT datasets. For the Commonsense dataset, we use the average accuracy across eight downstream tasks.

\section{Results}\label{sec:results}
\subsection{Editing Performance after Fine-tuning}\label{sec:results_ed_ft}
Tab.~\ref {tab:edit_perf_ft_wise} presents the ES of GPT-J, GPT2-XL, and Llama2 on \zsre, before and after fine-tuning on Commonsense datasets. Results for PS and NS, as well as the results for other experimental configurations (i.e., \cf and HotpotQA datasets, MEND method, and Llama 3.1 model), are provided in Tab.~\ref{tab:over_ke_perf_mend} and Tab.~\ref{tab:ds_perf_another} in App.~\ref{app:supplementary_results}. We also present the performance of KE-only ($M_{\text{ed}}$) and FT-only ($M_{\text{ft}}$) in App.~\ref{app:ke_perf_breakdown} and App.~\ref{app:ft_perf_breakdown}, validating the setup.

Overall, \textbf{fine-tuning reduces editing performance}, with only four \textcolor{darkpink}{exceptions} showing comparable editing performance to $M_\text{ed}$ (``No ft'' in Tab.~\ref{tab:edit_perf_ft_wise}). For example, when editing 1000 facts using MEMIT on GPT2-XL, after full fine-tuning, the ES increases by $3.37$ percentage points (p.p.). It is counterintuitive that fine-tuning on data unrelated to the target editing knowledge causes previously unsuccessful edits to succeed. We refer to such edits as \textit{Emergent edits} and analyze them further in Sec.~\ref{sec:qualitative_analysis}.
Among all \textcolor{darkgreen}{decay cases}, LoRA on Llama2 after 1000 edits on \zsre using AlphaEdit yields the largest decrease in editing performance, from $93.23\%$ to $50.45\%$. Additionally, while fine-tuning impairs KE performance, the extent of this effect varies across KE setups, fine-tuning configurations, and models, as discussed below.

\input{tabs/edit_perf_ft_wise}

\paragraph{Impact of FT configuration.}
From Fig.~\ref{fig:edit_perf_llama2}, we find that \textbf{full fine-tuning impairs a markedly larger fraction of edits than LoRA and DoRA}, also shown by the average decrease across all models in Tab.~\ref{tab:decrease_rate_after_ft}: $38.10\%$ for full fine-tuning, versus $28.71\%$ and $29.88\%$ for LoRA and DoRA. DoRA demonstrates a slightly stronger ability to remove edits than LoRA. This pattern varies with model architecture and edit scale. Besides, \textbf{models fine-tuned by HotpotQA exhibit larger decays compared to those being fine-tuned using Commonsense} (results and discussions are in App.~\ref{app:ke_after_hotpotqa}).

\paragraph{Impact of KE method.}
Between AlphaEdit and MEMIT, \textbf{AlphaEdit exhibits greater decay after FT}, i.e., its edits are more easily removed by FT. For example, in the left subplot of Fig.~\ref{fig:edit_perf_llama2}, LoRA reduces MEMIT (100 edits) performance on Llama2 by $9.73$ p.p., compared to $35.37$ p.p. for AlphaEdit, showing a noticeable gap of $25.64$ p.p. When the number of edits increases to $10{,}000$, this gap widens to $33.50$ p.p., indicating that large-scale edits exacerbate AlphaEdit’s vulnerability. Such a pattern is observed consistently across GPT-J and GPT2-XL. This phenomenon may stem from the \textit{null-space vulnerability} of AlphaEdit that the null space, which stores AlphaEdit's edits, is highly sensitive to fine-tuning. This is because the null space is nearly orthogonal to the existing knowledge space, its representation does not affect the model’s outputs. That is, the representation of null space is not used to compute the loss during fine‑tuning~\citep{fang2025alphaeditnullspaceconstrainedknowledge}. As a result, fine-tuning can freely introduce new or conflicting information into this space without penalty, thereby disrupting AlphaEdit’s stored knowledge. MEND‑edited models exhibit consistent patterns observed in other KE methods: as the number of edits increases, fine‑tuning removes a larger proportion of applied edits, as shown in Tab.~\ref{tab:over_ke_perf_mend}. Prior studies indicate that MEND, which adjusts existing weights based on training data, generally performs poorly in such zero‑shot–style settings \citep{fang2025alphaeditnullspaceconstrainedknowledge, zhang2024fisher}.

\definecolor{mygray}{RGB}{169,169,169}
\definecolor{darkblue}{RGB}{55, 103, 149}
\definecolor{orange}{RGB}{255, 208, 111}
\definecolor{reddishbrown}{RGB}{231, 98, 84}
\begin{figure*}[t]
    \centering
    \includegraphics[width=0.28\textwidth]{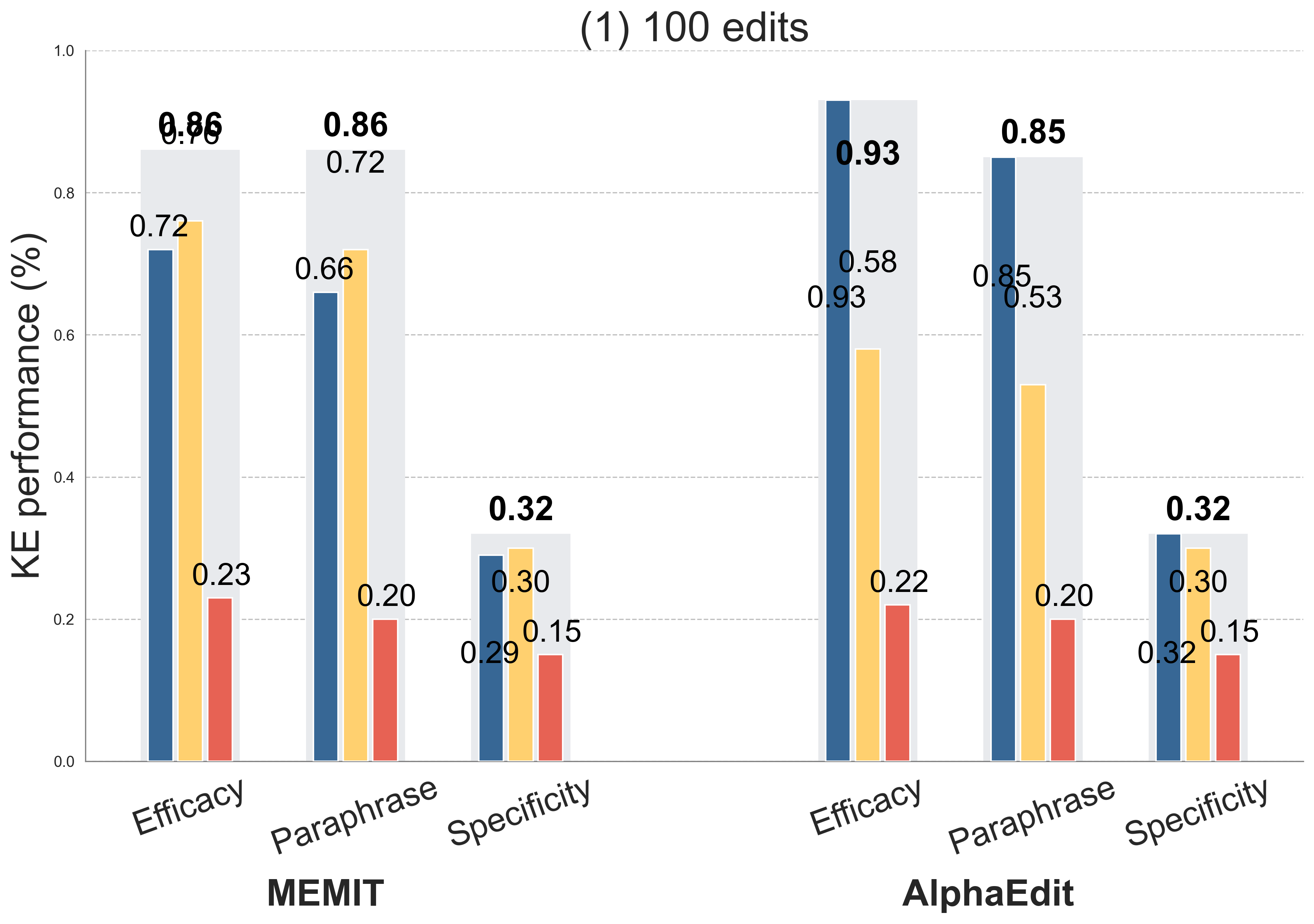}\hfill
    \includegraphics[width=0.28\textwidth]{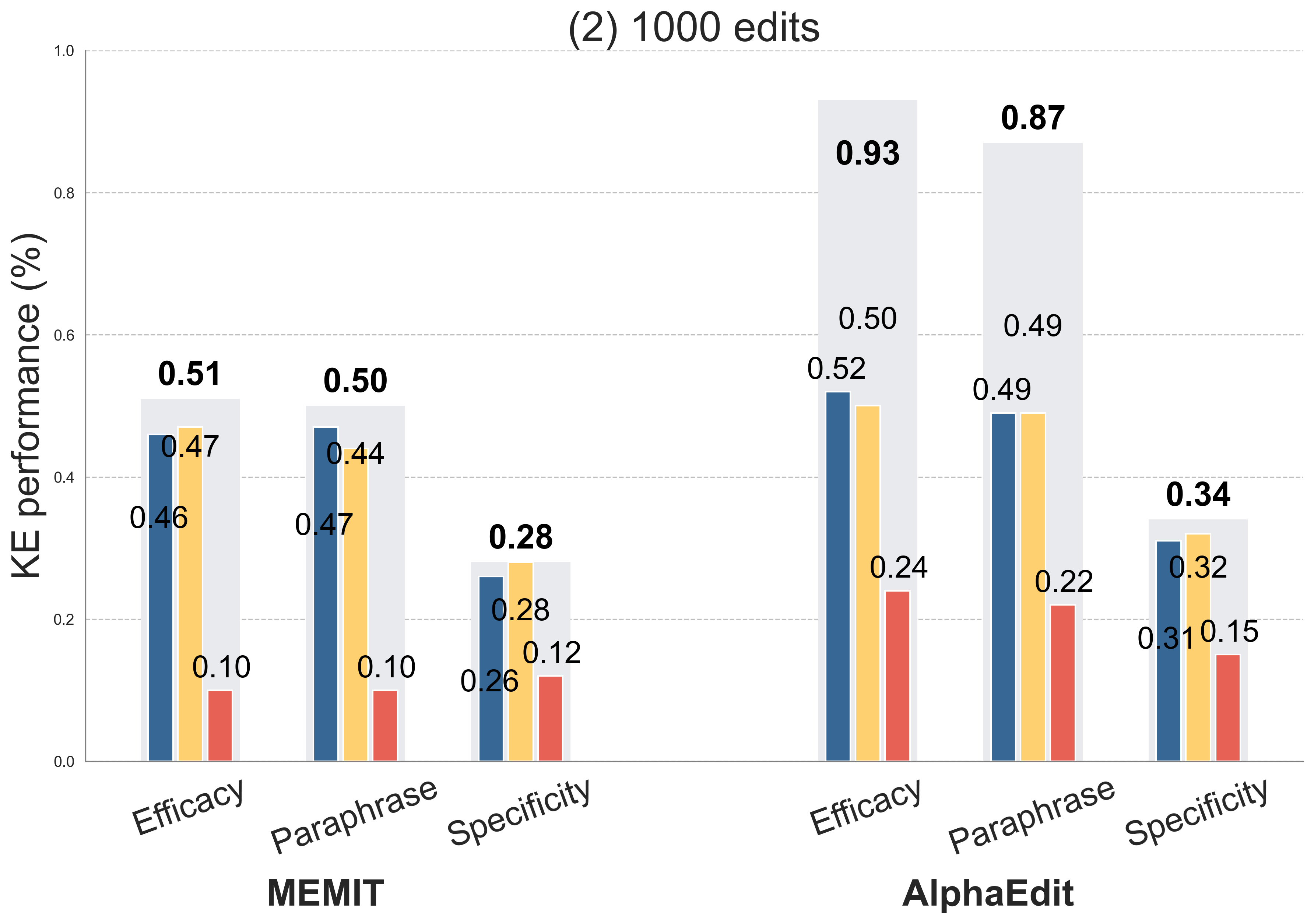}\hfill
    \includegraphics[width=0.28\textwidth]{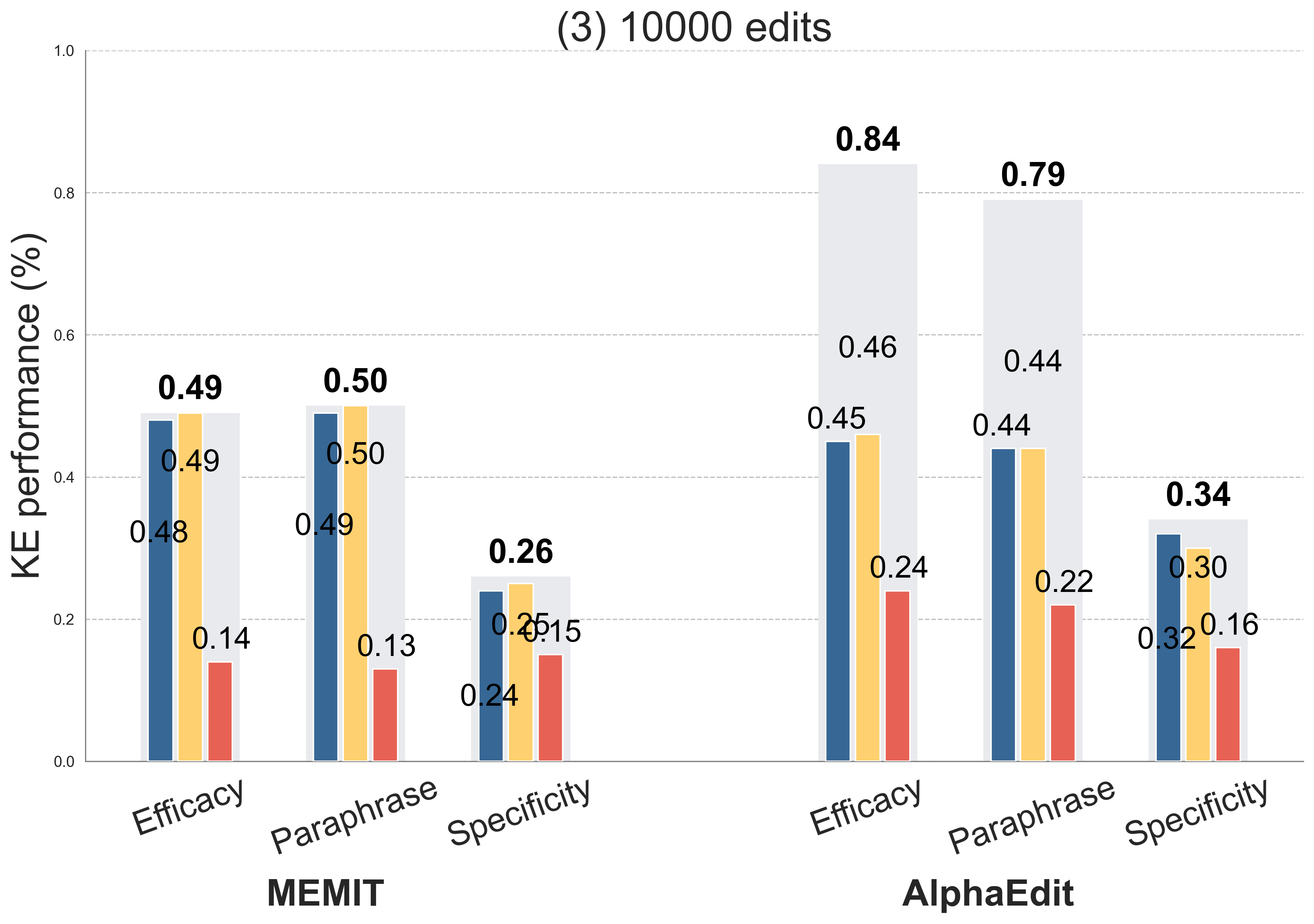}\\[2pt]
    \includegraphics[width=0.7\textwidth]{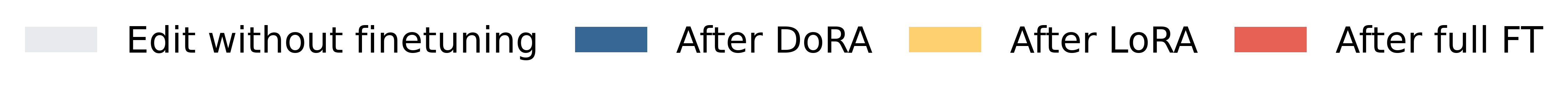}
    \vspace{-5pt}
    \caption{Editing performance of Llama2 on \zsre dataset before and after fine-tuning. Editing performance after fine-tuning ({\color{orange}LoRA}, {\color{darkblue}DoRA} and {\color{reddishbrown}full size fine-tuning}) is compared against the editing performance {\color{gray}before fine-tuning}.}
    \label{fig:edit_perf_llama2}
\end{figure*}

\paragraph{Impact of KE dataset.}
\input{tabs/app/over_ke_perf_mend}

\input{tabs/decrease_rate_after_ft}

As illustrated in Tab.~\ref{tab:edit_perf_ft_wise}, \textbf{models edited with \zsre generally experience a greater decline in KE performance compared to those edited with \cf}, regardless of the experimental configurations. For example, when GPT-J is MEMIT edited $10{,}000$ facts and LoRA fine-tuned, \zsre-edited one decreases $25.35$ p.p. more than \cf-edited one ($30.11-4.76$). The larger performance drop on \zsre likely stems from differences in evaluation metrics: \cf deems an edit successful if the edited object has a higher probability than the original, whereas \zsre requires the model to generate the edited object correctly, a stricter criterion.

\paragraph{Impact of Model.} 
As shown by the~\textit{Overall Avg.} in Tab.~\ref{tab:decrease_rate_after_ft}, \textbf{GPT-J is the most stable under fine-tuning, followed by Llama2, whereas GPT2-XL the largest variability}. GPT-J achieves the smallest average decrease ($15.93\%$) compared to Llama2 ($39.92\%$) and GPT2-XL ($40.84\%$). Llama3.1's performance is similar to Llama2 (see App.~\ref{app:ke_perf_breakdown}). This might be due to GPT-J having a larger MLP projection matrix which could lead to a higher editing robustness. Specifically, GPT-J utilizes a projection matrix of $16384 \times 4096$, surpassing the $14336 \times 4096$ dimensions of Llama2. Furthermore, since all models exhibit edit decay effects to varying degrees, we expect our findings to generalize to other LLMs, as modern systems share the same transformer architecture with only minor variations.

\paragraph{Edit Flip Ratio.}

As shown in Tab.~\ref{tab:efr_gptj}, for \zsre, EFR and $\Delta$ES exhibit similar trends, suggesting that fine-tuning primarily introduces flipping cases that account for the observed drop in efficacy. However, for some instances on \cf, such as \aedit with an edits of $10^4$, \textbf{EFR exceeds $\Delta$ES}, indicating the presence of reverse flipping cases (from fail to success) that offset some of the flipping cases and thus reduce the overall $\Delta$ES. In other words, a small change in ES does not necessarily reflect a small number of flipping edits. These cases are further discussed in Sec.~\ref{sec:qualitative_analysis}.

\input{tabs/efr_gptj}

\subsection{Qualitative Analysis}\label{sec:qualitative_analysis}
We conduct a qualitative analysis by manually examining examples of different model behaviors on target edits after fine-tuning. Tab.~\ref{tab:quality_analysis_examples} summarizes several identified patterns. First, we observe stable cases where successful edits persist after fine-tuning. These \emph{Stable Edits} typically involve frequent lexical items as the targets, such as ``English'', ``Islam'', and ``piano''. By contrast, \emph{Erased Edits}, where the updated knowledge is removed after fine-tuning, tend to involve less frequent terms (i.e., ``Lecanorales''), suggesting that frequency and entrenchment of the target knowledge potentially influence the stability of edits. We also observe this pattern in \textit{Emergent Edits} where unsuccessful edits become successful ones after FT, wherein the target knowledge involves high-frequency tokens. For instance, when querying Danielle Darrieux’s mother tongue, $M_{\text{ed}}$ should output English but instead returns ``United States''. After FT, however, $M_{\text{ed\_ft}}$ produces ``English'', a frequent word. We attribute this emergence to conflicts between edits: in batch editing~\citep{meng2023masseditingmemorytransformer}, one edit may increase the probability of ``English'' while lowering the probabilities of other languages (e.g., “French”), whereas another edit may simultaneously suppress ``English'' as a side effect of optimizing a different target (e.g., ``Russian''). This tension renders some edits nonfunctional in $M_{\text{ed}}$. By erasing conflicting edits, FT inadvertently resolves this tension, allowing the previously suppressed edit to become functional in $M_{\text{ed\_ft}}$.

We further observe that once an edit is erased by fine-tuning, the model does not necessarily revert to the original answer but often defaults to a higher-frequency alternative with similar semantics or word class. For example, in the third case of \textit{Erased Edits}, after the target ``Philadelphia'' is removed, $M_{\text{ed\_ft}}$ outputs ``London'' rather than the original answer ``Paris''. Further, we find an interesting case that in the final case of \textit{Impossible Edits}, both $M_{\text{ed}}$ and $M_{\text{ed\_ft}}$ return ``1 May 1977'', whereas the expected answer is ``12 May 1977''. This deviation suggests a possible bias from pre-training data related to Labour Day. Furthermore, Liu’s work~\cite{liu2025modeleditingbuiltsand} suggests that existing KE methods often prioritize keyword substitution over genuine semantic modification, potentially explaining the editing patterns observed above. We leave this to future investigation.

\input{tabs/quality_analysis_examples}

\subsection{Only Fine-tuning Edited or Non-Edited Layers}\label{sec:only_ft_edlayers}
As discussed in Sec.~\ref{related_work_safety}, 
one of our motivations is to understand how to remove harmful edits and preserve beneficial ones. Meanwhile, as shown in~\ref{sec:results_ed_ft}, FT removes edits from $M_\text{ed}$. Taken together, we propose two hypotheses:
(1) FT only edited layers (denote as $M_{ed\_ft\_edited}$) can effectively remove edits;
(2) FT non-edited layers (denote as $M_{ed\_ft\_non-edited}$) can preserve edits. 
Besides, we denote the control group, fine-tuning with all layers, as $M_{ed\_ft\_all}$.

To examine this, we set two experimental groups: fine-tuning only the edited layers and fine-tuning only the non-edited layers. For our experiments, we adopt Llama2 and GPT-J as base models (test configurations along with results breakdown are in App.~\ref{app:only_ft_selected_layers}).

\input{tabs/ft_selected_layers_keperf}

\paragraph{Fine-tuning only edited layers.}
First, we find that fine-tuning only the edited layers can remove more prior edits than fine-tuning all layers. As illustrated in Tab.~\ref{tab:ft_selected_layers_keperf}, in the case of 100 Edits, between $M_\text{ed\_ft\_all}$ and $M_\text{ed\_ft\_edited}$, $M_\text{ed\_ft\_edited}$ shows a larger drop of editing performance across all KE metrics. For example, ES of $M_\text{ed\_ft\_edited}$ drops to $66$\% while ES of $M_\text{ed\_ft\_all}$ rises to $98$\%, close to $96$\% of $M_\text{ed}$. However, in the bottom of Tab.~\ref{tab:ft_selected_layers_keperf}, \textit{DS} of $M_\text{ed\_ft\_all}$ is larger than that of $M_\text{ed\_ft\_non-edited}$ or $M_\text{ed\_ft\_edited}$, indicating that fine-tuning only edited layers can result in a loss of downstream performance (see Tab.~\ref{tab:ft_selected_layers_keperf}).

This underscores the trade-off between the effectiveness of edit removing and the degradation of downstream performance. If overall downstream performance is not a priority, fine-tuning only the edited layers is an effective strategy for removing unwanted edits. 

\paragraph{Fine-tuning only non-edited layers.}
After comparing $M_{ed\_ft\_all}$ with $M_{ed\_ft\_non-edited}$ in Tab.~\ref{tab:ft_selected_layers_keperf}, we find a counterintuitive result for hypothesis~$(2)$: \textbf{fine-tuning only non-edited layers does not help preserve edits}. For example, with 100 \aedit-edits on Llama2, $M_{ed\_ft\_non-edited}$ shows a significant decline in ES from 98\% to 72\%. A similar trend is observed in PS, where $M_{ed\_ft\_non-edited}$ exhibiting greater degradation than $M_{ed\_ft\_all}$. We further investigate whether fine-tuning only the non-edited layers can be an edit-removal strategy (similar to $M_{ed\_ft\_edited}$'s performance). As shown in Tab.~\ref{tab:ft_selected_layers_keperf}, this approach maintains stronger downstream performance (80.61 vs. 65.43; 72.46 vs. 65.35, all in \%) while erasing fewer edits compared to fine-tuning only the edited layers. These results suggest that fine-tuning non-edited layers can be a supplementary edit-removal strategy.

\paragraph{Discussion.}
With respect to edits removal, we observe the following ordering across different settings: 

$M_{ed\_ft\_edited} \allowbreak > \allowbreak
 M_{ed\_ft\_non-edited} \allowbreak > \allowbreak
 M_{ed\_ft\_all}$. While for downstream task performance, we have the following order: $M_{ed\_ft\_all} \allowbreak > \allowbreak
 M_{ed\_ft\_non-edited} \allowbreak > \allowbreak
 M_{ed\_ft\_edited}$. 
 
These observations align with the \textbf{distributed representation hypothesis}, which posits that factual associations in LLMs are associated with multiple MLP and attention layers \citep{geva-etal-2023-dissecting, dar-etal-2023-analyzing}. KE adjusts weights and stores knowledge across multiple layers, whereas fine-tuning can readily disrupt this coordinated structure that supports these edits. Adapting fewer layers leads to larger weight updates within each layer, causing more substantial disruption to the stored edits. This explains why $M_\text{ed\_ft\_edited}$ and $M_\text{ed\_ft\_non-edited}$ erases more edits than $M_\text{ed\_ft\_all}$. Moreover, because the layers identified by Causal Tracing are most influential for editing knowledge~\citep{meng2023locatingeditingfactualassociations}, fine-tuning only these layers is intuitively more effective at removing edits, leading $M_\text{ed\_ft\_edited}$ to outperform $M_\text{ed\_ft\_non-edited}$ in edit removal.

\subsection{Impact of KE on post‑FT downstream performance}\label{sec:reasoning_ability_check}
\input{tabs/decrease_rate_caused_by_KE}
To evaluate how KE influences subsequent fine-tuning, we compare downstream performance across $M_{\text{ed}}$ and its fine-tuned counterpart $M_{\text{ed\_ft}}$. We find that KE moderately reduces the effectiveness of fine-tuning, with the degree of impact contingent upon specific fine-tuning methods and editing parameters (see App.~\ref{app:ke_perf_breakdown}). As shown in Tab.~\ref{tab:decrease_rate_caused_by_KE}, we found FT methods influence KE robustness differently: regarding FT methods, where LoRA achieves the smallest average performance degradation while DoRA provides the most consistent stability across models; regarding FT models, GPT2-XL demonstrates the highest robustness to KE across different methods and datasets, whereas other models such as Llama2 exhibit substantially larger performance fluctuations; regarding KE methods, MEMIT generally preserves fine-tuning performance better than AlphaEdit across models and datasets. Furthermore, our analysis indicates that \textbf{downstream performance degradation is not attributable to catastrophic forgetting}, as $M_\text{ed\_ft}$ maintains comparable 
downstream performance to $M_\text{ft}$ while outperforming the $M_\text{ed}$.

\section{Why Editing is Fragile to Subsequent Fine-tuning}\label{sec:act_analysis_main}

\definecolor{darkblue2}{RGB}{0, 114, 178}
\definecolor{orange2}{RGB}{213,94,0}
\definecolor{cyan2}{RGB}{0,158,115}
\definecolor{vermillion}{RGB}{213, 94, 0}
\definecolor{skyblue}{RGB}{86, 180, 233}
\definecolor{purple}{RGB}{204, 121, 167}
\begin{figure*}[t]
    \centering
    \includegraphics[width=0.25\textwidth]{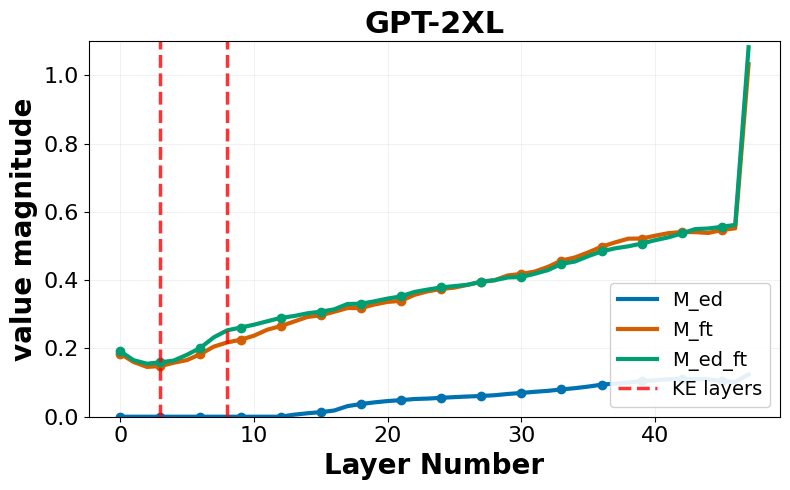}\hspace{10pt}
    \includegraphics[width=0.25\textwidth]{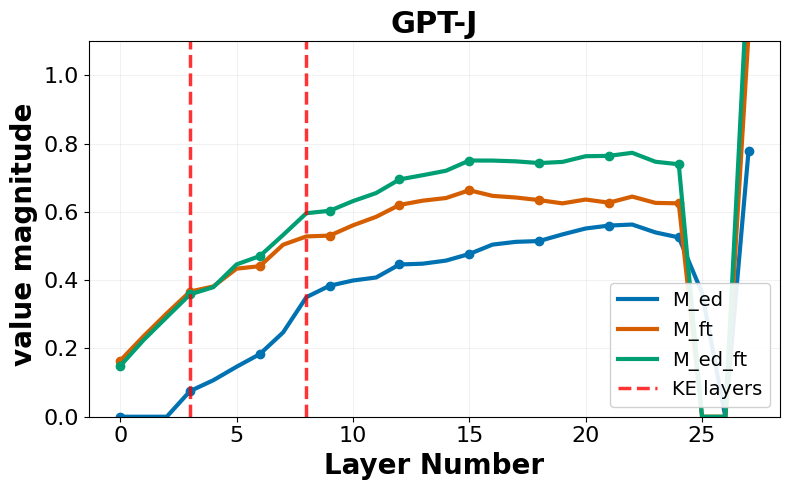}\hspace{9pt}
    \includegraphics[width=0.25\textwidth]{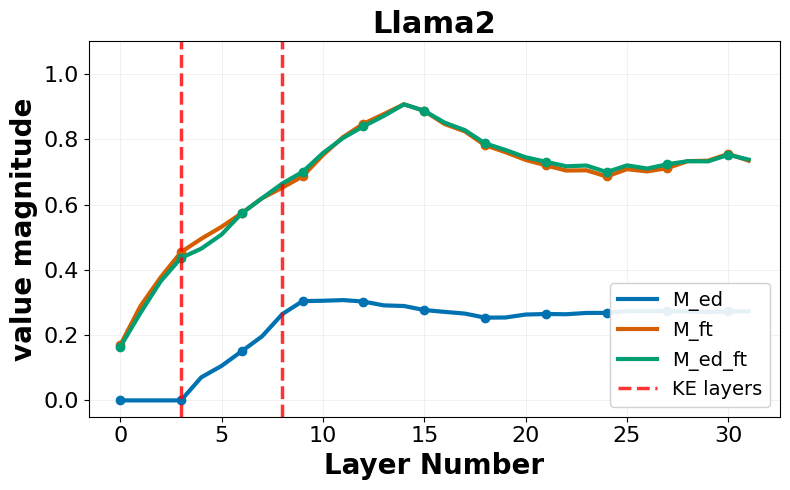} \\
    \includegraphics[width=0.25\textwidth]{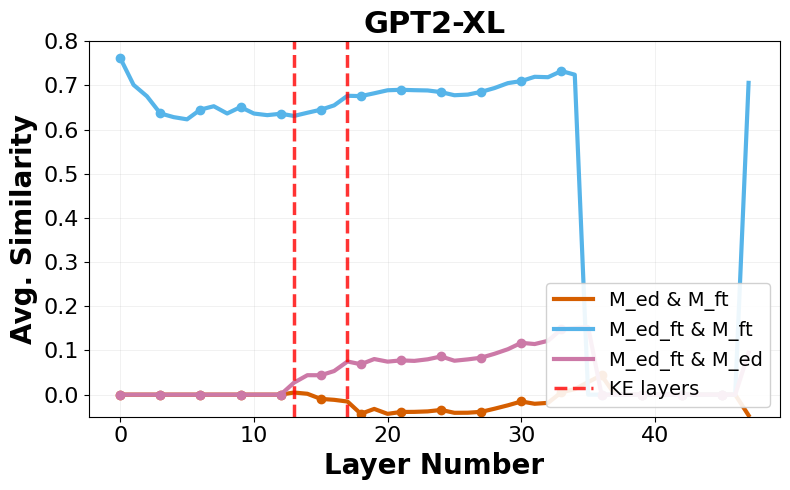}\hspace{10pt}
    \includegraphics[width=0.25\textwidth]{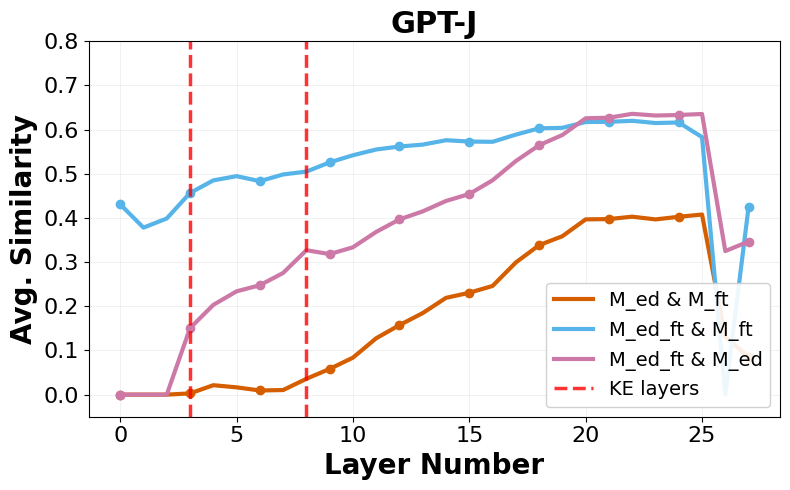}\hspace{10pt}
    \includegraphics[width=0.25\textwidth]{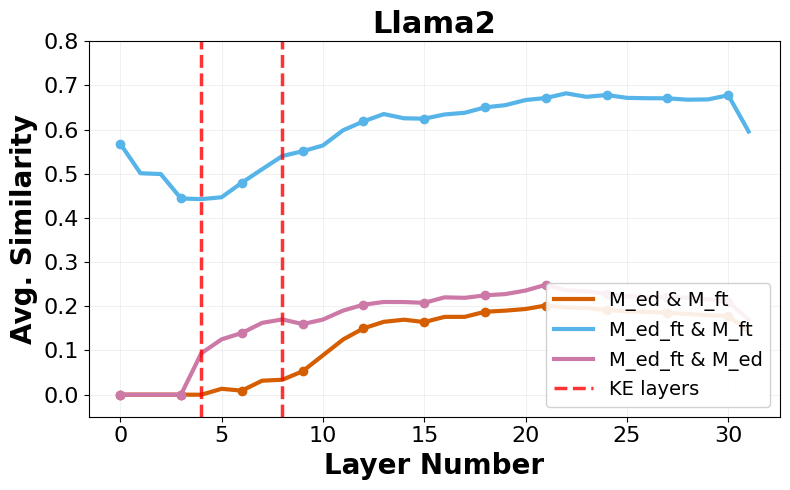}
    \caption{In \textbf{layer-wise activation drifts} (top row) for GPT2-XL, GPT-J and Llama2, 3 categories for each model: {\color{darkblue2}$M_\text{ed}$}, {\color{orange2}$M_\text{ft}$} and {\color{cyan2}$M_\text{ed\_ft}$}. In \textbf{directional similarities}  (bottom row), 3 pairs of categories tested for each model: {\color{vermillion}$M_\text{ed}$ - $M_\text{ft}$}, {\color{skyblue}$M_\text{ed\_ft}$ - $M_\text{ft}$} and {\color{purple}$M_\text{ed\_ft}$ - $M_\text{ed}$}. Within the red vertical dash lines are the range of layers being edited. Result specifications in App.~\ref{app:activation_res_breakdown}}
    \label{fig:layer_drift}
\end{figure*}

To understand why editing is fragile to subsequent fine-tuning, we investigate whether a KE induces a coherent, structured shift in the activation space and how subsequent fine-tuning perturbs or erodes this shift. 
For a model \(M\), its edited version \(M_{\text{ed}}\), and its edited-then-fine-tuned version \(M_{\text{ed\_ft}}\), we analyze activations \(h_\ell(x)\) at each layer \(\ell\) using prompt $x$ from a diagnostic prompt set \(\mathcal{X}\), which comprises two groups: prompts that explicitly query the edited knowledge (from the editing dataset, Sec.~\ref{sec:editing_datasets})  and prompts that do not directly invoke the edited fact (from downstream tasks, Sec.~\ref{sec:fine-tune_datasets}). Each group has 40 prompts that are shuffled after sampling from the two sources.

\subsection{Layer-wise magnitude changes}\label{sec:layer_wise_mag}

For each prompt \(x\) in \(\mathcal{X}\), we compute the magnitude of activation changes introduced by editing (\(\Delta^{\text{ed}}_\ell(x)\)), by fine-tuning (\(\Delta^{\text{ft}}_\ell(x)\)) and by both (\(\Delta^{\text{ed\_ft}}_\ell(x)\)),  where :
{
\scriptsize 
\begin{equation}
\begin{aligned}
\Delta^{\text{ed}}_\ell(x) &= \| h_\ell^{M_\text{ed}}(x) - h^M_\ell(x) \|_2 \\
\Delta^{\text{ft}}_\ell(x) &= \| h_\ell^{M_\text{ft}}(x) - h^M_\ell(x) \|_2 \\
\Delta^{\text{ed\_ft}}_\ell(x) &= \| h_\ell^{M_\text{ed\_ft}}(x) - h_\ell^M(x) \|_2
\end{aligned}
\label{eq:actsps_shift} 
\end{equation} 
}

\definecolor{darkblue2}{RGB}{0, 114, 178}
\definecolor{orange2}{RGB}{213,94,0}
\definecolor{cyan2}{RGB}{0,158,115}
\definecolor{vermillion}{RGB}{213, 94, 0}
\definecolor{skyblue}{RGB}{86, 180, 233}
\definecolor{purple}{RGB}{204, 121, 167}
\begin{figure*}
    \centering
    \includegraphics[width=0.25\textwidth]{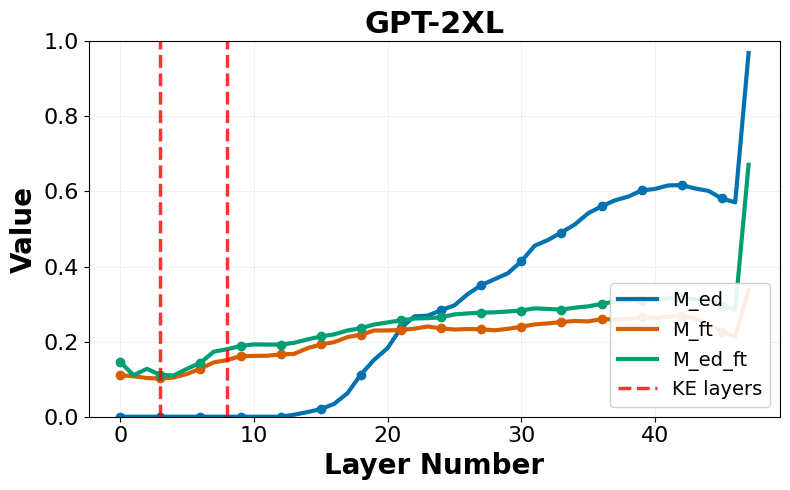}\hspace{10pt}
    \includegraphics[width=0.25\textwidth]{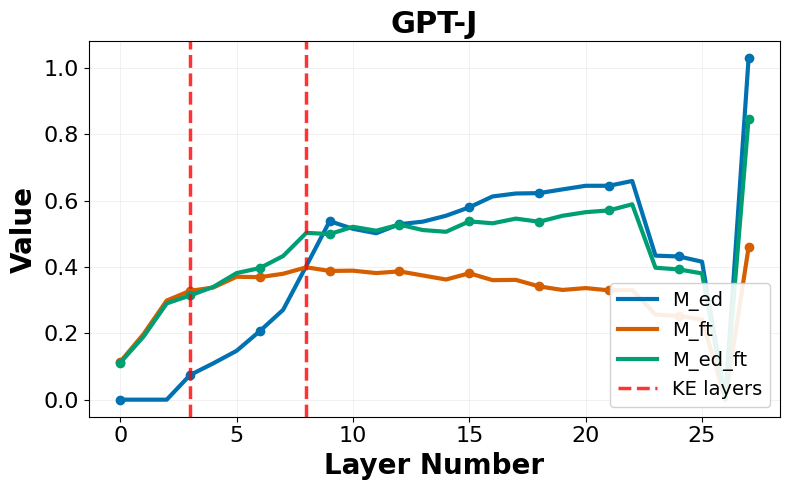}\hspace{10pt}
    \includegraphics[width=0.25\textwidth]{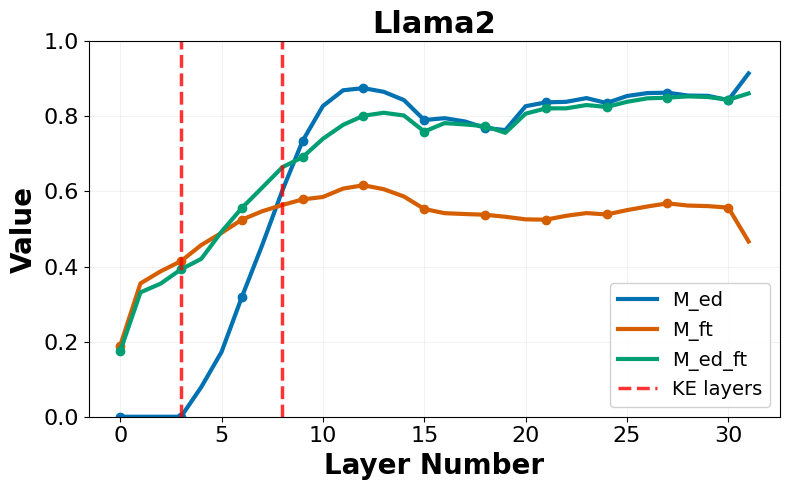} \\
    \includegraphics[width=0.25\textwidth]{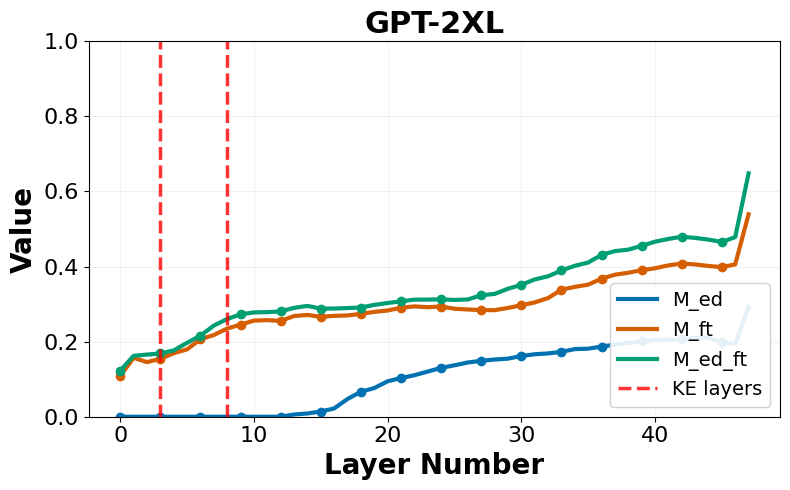}\hspace{10pt}
    \includegraphics[width=0.25\textwidth]{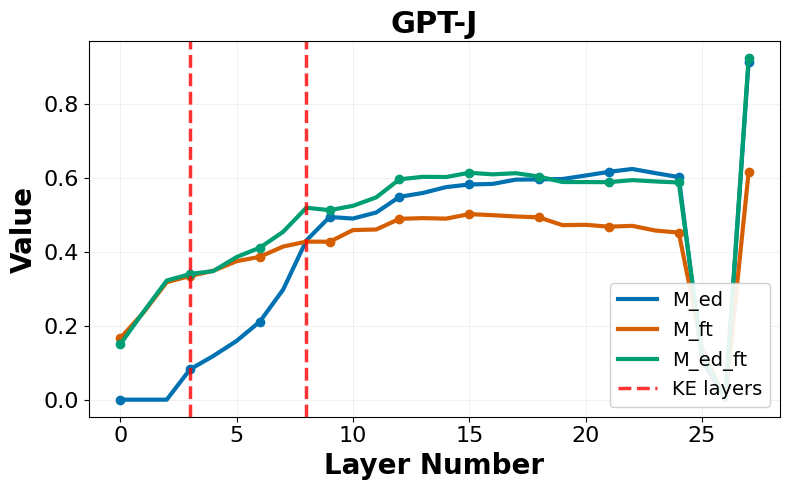}\hspace{10pt}
    \includegraphics[width=0.25\textwidth]{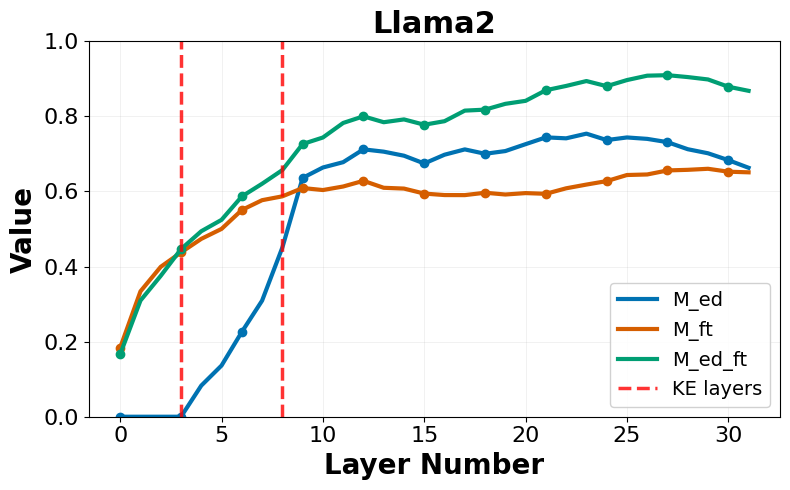}
    \caption{Layer-wise activation magnitude change for GPT2-XL, GPT-J and Llama2, 3 categories for each model: {\color{darkblue2}$M_\text{ed}$}, {\color{orange2}$M_\text{ft}$} and {\color{cyan2}$M_\text{ed\_ft}$}; 2 edit types: stable edits (top row), erased edits (bottom row).}
    \label{fig:layer_drift_cases}
\end{figure*}

\begin{figure*}
    \centering
    \includegraphics[width=0.25\textwidth]{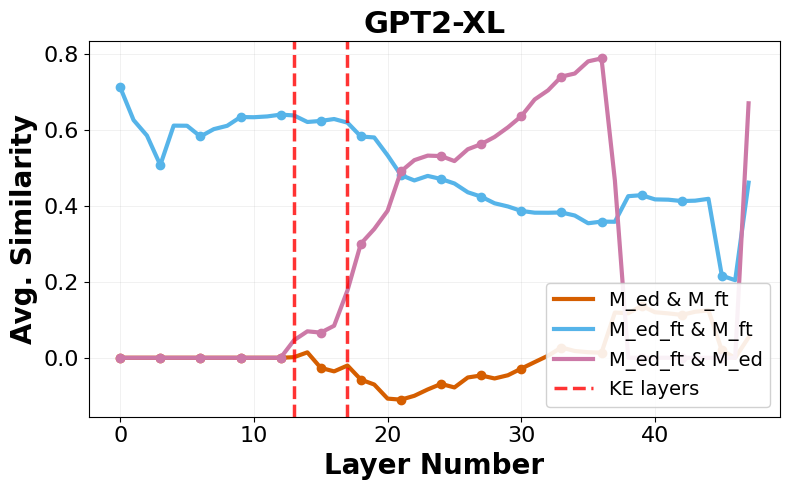}\hspace{10pt}
    \includegraphics[width=0.25\textwidth]{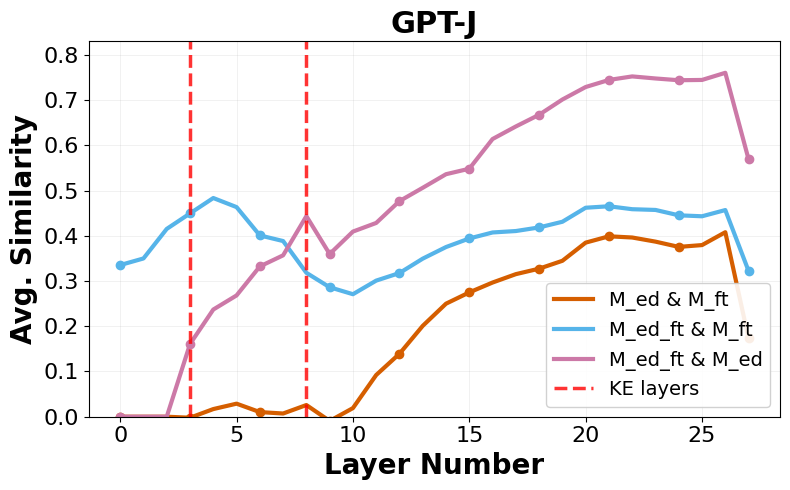}\hspace{10pt}
    \includegraphics[width=0.25\textwidth]{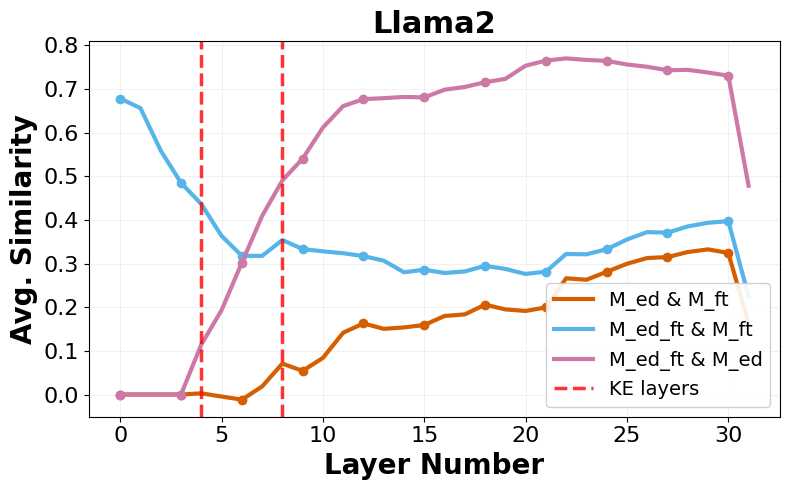} \\
    \includegraphics[width=0.25\textwidth]{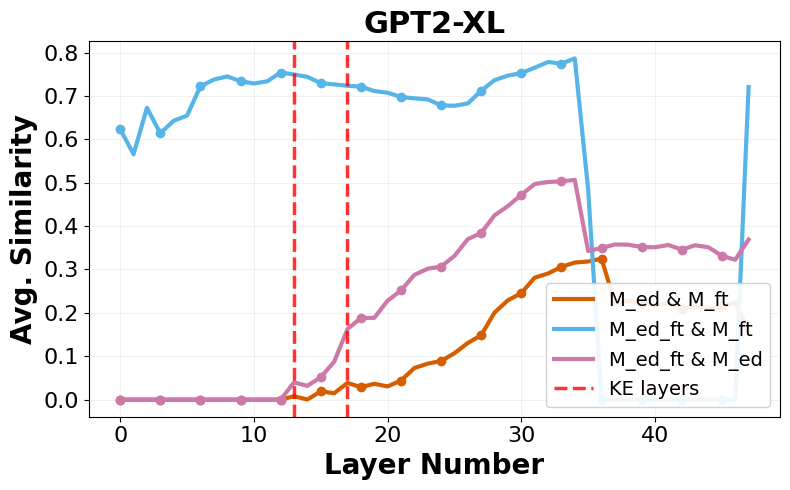}\hspace{10pt}
    \includegraphics[width=0.25\textwidth]{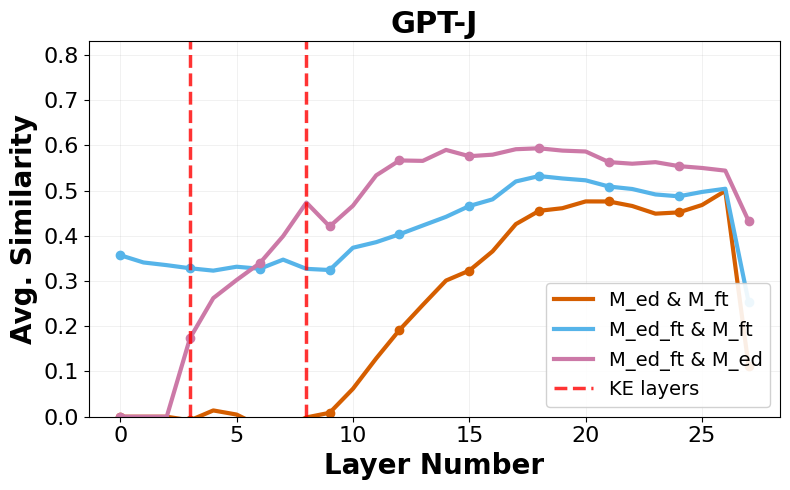}\hspace{10pt}
    \includegraphics[width=0.25\textwidth]{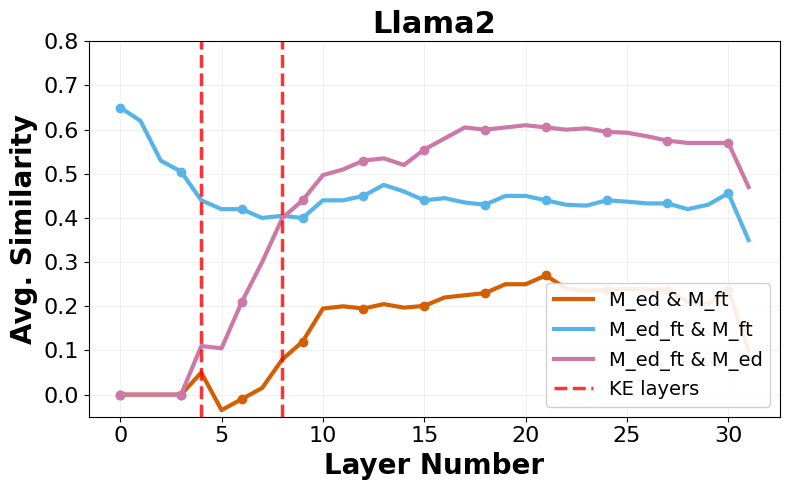}
    \caption{Layer-wise activation directional similarity for GPT2-XL, GPT-J and Llama2, 3 pairs of categories tested for each model: {\color{vermillion}$M_\text{ed}$ - $M_\text{ft}$}, {\color{skyblue}$M_\text{ed\_ft}$ - $M_\text{ft}$} and {\color{purple}$M_\text{ed\_ft}$ - $M_\text{ed}$}; 2 edit types: stable edits (top row), erased edits (bottom row).}
    \label{fig:dir_sim_cases}
\end{figure*}

We compute the arithmetic mean of activation deltas over all prompts in \(\mathcal{X}\) and visualize activation changes across layers in the first row of Fig.~\ref{fig:layer_drift}. We make two key observations. First, fine-tuned models, both {\color{orange2}{$M_\text{ft}$}} and {\color{cyan2}{$M_\text{ed\_ft}$}} in the first row of Fig.~\ref{fig:layer_drift}, exhibit substantially larger magnitude changes than their non-fine-tuned counterparts across all layers; they also show similar magnitudes. This separation is particularly pronounced for GPT-2XL and Llama2, where a consistent gap between fine-tuned and non-fine-tuned models is observed at every layer. Second, edited-only models present major changes in layers from the edited layer, while fine-tuning affects a broader range of layers. 
These findings suggest that edits induce limited and localized activation perturbations, which can be overwritten by the global representational drifts induced by fine-tuning, consistent with the degradation effects reported in Sec.~\ref{sec:only_ft_edlayers}.

\subsection{Layer-wise directional drift}

To further characterize how fine-tuning interacts with KE regarding directions in representative space, we quantify the alignment between activation displacements induced by KE and by subsequent FT via computing the cosine similarity between activation displacements of $M_\text{ed}$ and $M_\text{ed\_ft}$ using the same prompt set \(\mathcal{X}\) above. We first define the layer‑wise displacement vector as the difference between the activations of a model and its un-edited/fine-tuned counterparts. Accordingly, the displacement vectors of {\color{vermillion}{$M_\text{ed}$}}, {\color{skyblue}{$M_\text{ft}$}} and {\color{purple}{$M_\text{ed\_ft}$}} are denoted by \(\Delta^{\text{ed}}_\ell(x)\), \(\Delta^{\text{ft}}_\ell(x)\) and \(\Delta^{\text{ed\_ft}}_\ell(x)\) respectively, as given in Eqn.~\ref{eq:actsps_dir1}. We then compute the layer-wise directional similarity by averaging similarities of all prompts in $\mathcal{X}$ at layer $\ell$, as shown in Equ.~\ref{eq:actsps_dir1} ($\varepsilon\!=\!10^{-8}$ is used to prevent division by zero):

{
\scriptsize
\begin{gather}
\Delta^{\text{ed}}_\ell(x) = h_\ell^{M_\text{ed}}(x) - h^M_\ell(x), \\
\Delta^{\text{ft}}_\ell(x) = h_\ell^{M_\text{ft}}(x) - h^M_\ell(x), \\ 
\Delta^{\text{ed\_ft}}_\ell(x) = h_\ell^{M_\text{ed\_ft}}(x) - h_\ell^{M_\text{ed}}(x). \\
\text{sim}_{\ell}(x) =\frac{1}{|\mathcal{X}|} \sum_{x\in\mathcal{X}}
\frac{\langle \Delta^{\text{1}}_{\ell}(x),\; \Delta^{\text{2}}_{\ell}(x)\rangle}
{\|\Delta^{M_\text{1}}_{\ell}(x)\|_{2}\; \|\Delta^{M_\text{2}}_{\ell}(x)\|_{2} + \varepsilon}.
\label{eq:actsps_dir1} 
\end{gather}
}

A value $\text{sim}_{\ell}\!\approx\!1$ indicates that fine-tuning pushes activations further in the direction of the KE, whereas a negative value suggests they are in the opposite direction. Intuitively, higher values indicate that knowledge editing and subsequent fine-tuning induce more aligned directional shifts in the activation space.

As shown in the bottom row of Fig.~\ref{fig:layer_drift}, across all three models, edited only model exhibits substantially lower similarity to the fine-tuned models than the similarity among the fine-tuned models themselves ({\color{vermillion}$M_\text{ed}$ - $M_\text{ft}$} $<$ {\color{purple}$M_\text{ed\_ft}$ - $M_\text{ed}$} $<<$ {\color{skyblue}$M_\text{ed\_ft}$ - $M_\text{ft}$}). This indicates that fine-tuning shifts activation in a direction that is \emph{nearly orthogonal to} the editing direction, which explains why the activations of $M_\text{ed}$ are least similar to those of $M_\text{ft}$.


We further examine whether the patterns discussed in the above paragraphs vary by editing types, i.e., separating stable edits and erased edits for layer-wise magnitude changes and directional drift analysis. 
For each type of edit, we randomly select 20 prompts from its corresponding pool of stable or erased edits. Each edit is classified by comparing the outputs of $M_{ed}$ and $M_{ed\_ft}$. An edit is considered \textit{stable} if both $M_{ed}$ and $M_{ed\_ft}$ give the correct answer, whereas it is regarded as an \textit{erased edit} if $M_{ed\_ft}$ gives the wrong answer. As illustrated in Fig.~\ref{fig:layer_drift_cases}, in terms of activation changes in magnitude, all {\color{darkblue2}edited models} (GPT2-XL, GPT-J, Llama2) exhibit larger activation shifts for stable edits (top row) compared to erased edits (bottom row). The gap narrows after FT, as the curves of {\color{orange2}$M_{ft}$} and {\color{cyan2}$M_{ed\_ft}$} are flatter than that of {\color{darkblue2}$M_{ed}$} curve. For directional drift (shown in Fig.~\ref{fig:dir_sim_cases}), models of erased edits show slightly higher similarity than those of stable edits, as the curves of {\color{vermillion}$M_{ed} - M_{ft}$} and {\color{purple}$M_{ed\_ft} - M_{ft}$} in the top row lie above their counterparts in the bottom row. 

To summarize, erased edits induce a smaller activation change in magnitude than stable edits during KE. Compared with stable edits, erased edits produce activations whose directional similarity to FT‑induced activations is higher. It may account for the fragile nature of these edits under FT.

\subsection{Geometric Shifts in Representation Space via UMAP}\label{sec:act_analysis_umap}

\definecolor{M}{RGB}{149,165,166}
\definecolor{M_ed}{RGB}{255,107,107}
\definecolor{M_ft}{RGB}{78,205,196}
\definecolor{M_ed_ft}{RGB}{255,230,109}
\begin{figure*}  
    \centering
    \includegraphics[width=0.23\textwidth]{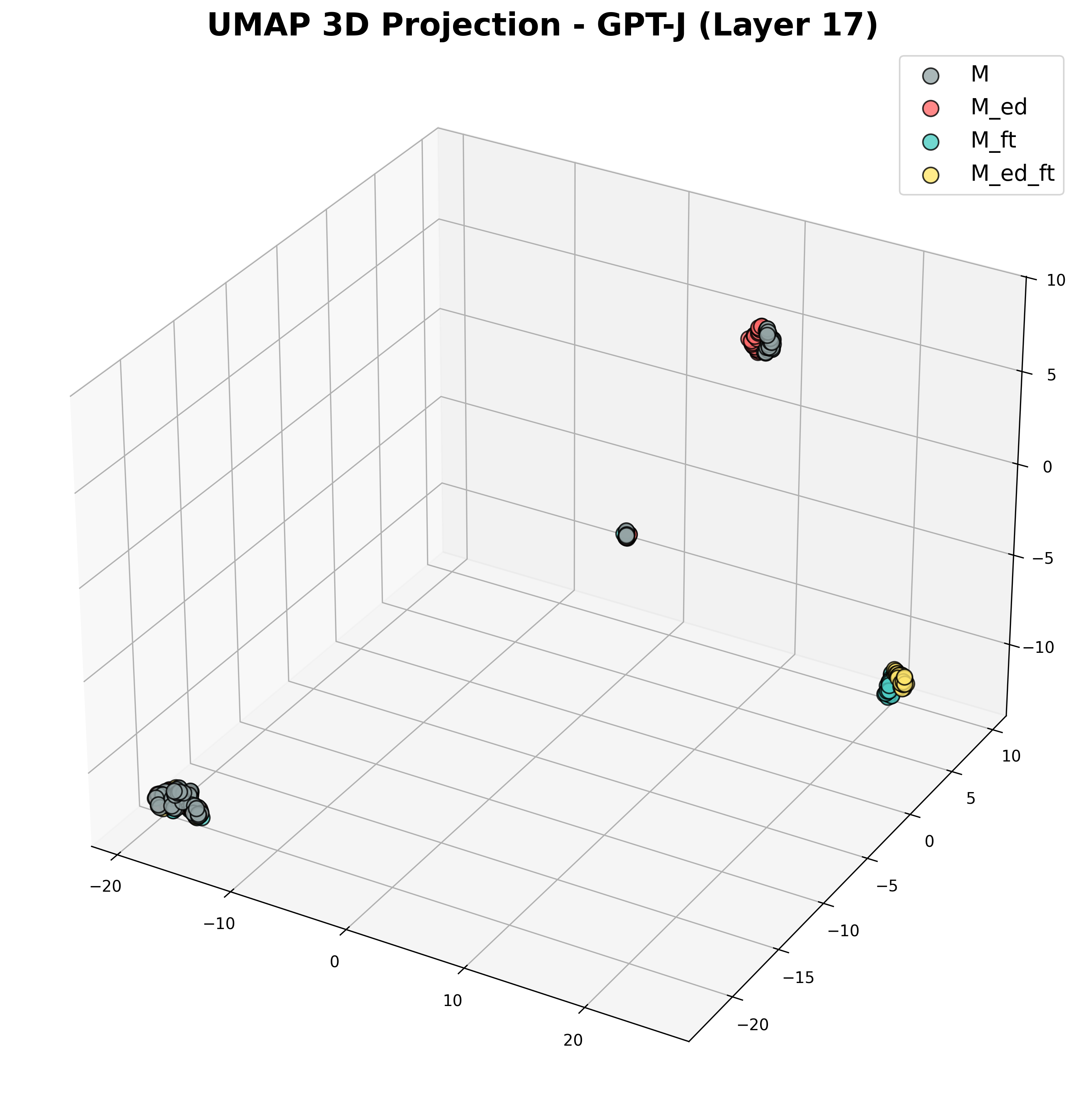}\hspace{20pt}
    \includegraphics[width=0.23\textwidth]{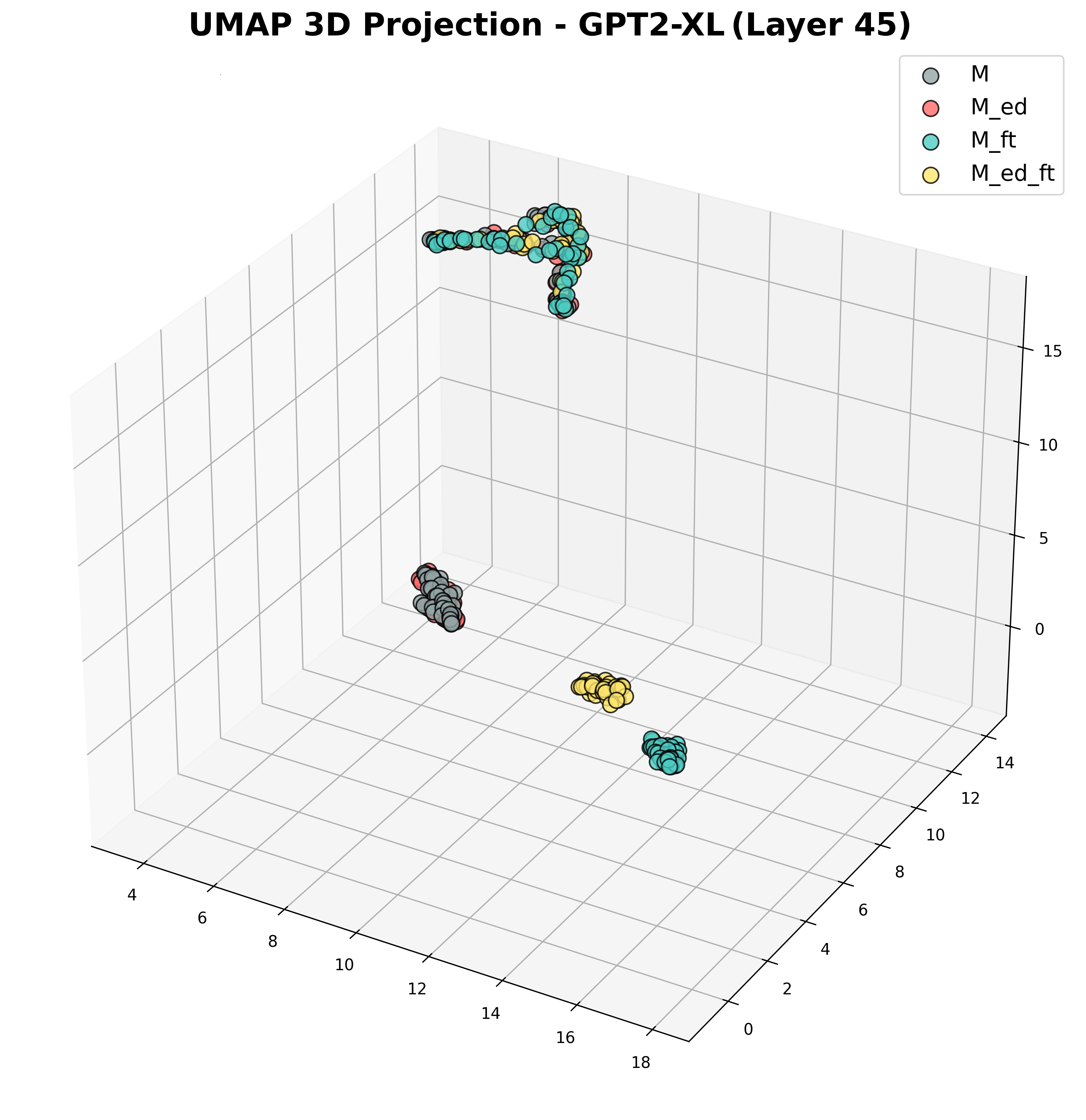}\hspace{20pt}
    \includegraphics[width=0.23\textwidth]{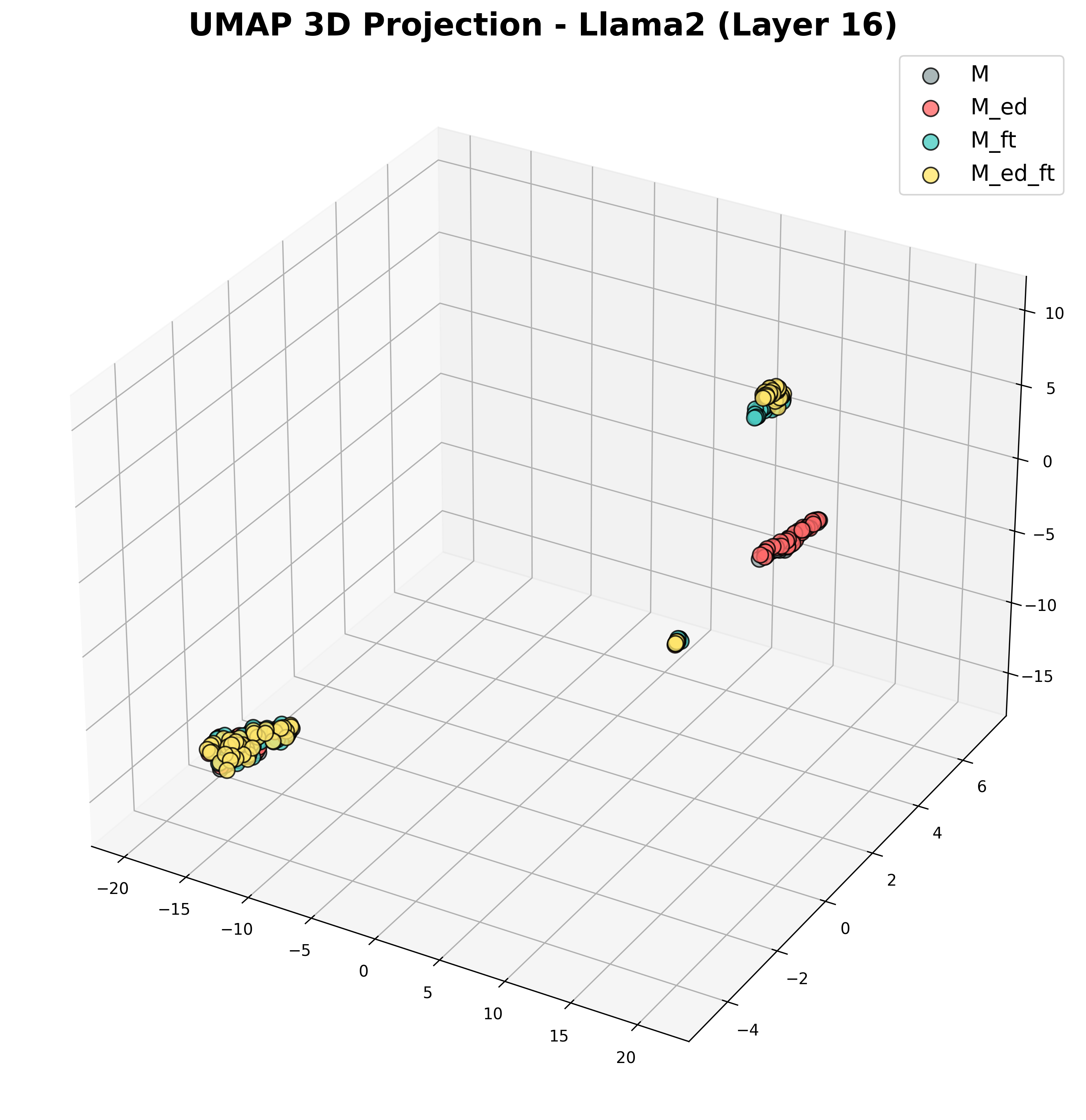}\\
    \caption{UMAP analysis of activations in 3D dimension. From left to right: GPT-J, GPT2-XL, Llama2.}
    \label{fig:act_umap}
\end{figure*}

In addition to the magnitude and direction changes, we further perform Uniform Manifold Approximation and Projection (UMAP) analysis~\citep{mcinnes2020umapuniformmanifoldapproximation} to study how the representations of various prompts are affected by KE and FT. UMAP projects high‑dimensional vectors into a lower‑dimensional space; its formal definition is given in Eqn.~\ref{eq:umap_def} in App.~\ref{app:activation_res_breakdown}. As inputs to UMAP, we use the activations of all layers for a given prompt $x$ in $\mathcal{X}$ across different model categories ($M$, $M_{ed}$, $M_{ft}$, and $M_{ed\_ft}$). Prompt set \(\mathcal{X}\) is identical to the one used for magnitude analysis in Sec.~\ref{sec:layer_wise_mag}. In each UMAP plot, we use the activation of one layer. The UMAP hyperparameters (20 neighbors and minimum distance of 0.1) are obtained via grid search to maximize cluster separation. 

We select the image of one post-KE layer from each model and the images of the rest layers are shown in App.~\ref{app:activation_res_breakdown}. As shown in Fig.~\ref{fig:act_umap}, across all models, the \textcolor{M_ft}{$M_{ft}$} cluster substantially overlaps with that of \textcolor{M_ed_ft}{$M_{ed\_ft}$}, while both are clearly separated from \textcolor{M_ed}{$M_{\text{ed}}$}. This pattern indicates that the outputs of \textcolor{M_ed_ft}{$M_{ed\_ft}$} are more similar to those of \textcolor{M_ft}{$M_{ft}$} compared to those of \textcolor{M_ed}{$M_{ed}$}. This suggests that fine-tuning effectively \textit{pulls} the activation of a KE-then-FT model (\textcolor{M_ed_ft}{$M_{ed\_ft}$}) towards fine-tuning's direction (\textcolor{M_ft}{$M_{ft}$}), causing it become more similar to the FT-only model, which might be leading to the removal of edits. The above observation also aligns with the findings of magnitude and directional analysis, which shows that fine-tuning has a larger impact on the activation than KE. Besides, \textcolor{M_ed}{$M_{ed}$}'s activations substantially overlap with with those of \textcolor{M}{$M$} in the representation space, suggesting that KE only induces relatively small changes in the model's internal representations. These small changes might be easily erasable when finetuning the model.

\begin{figure}  
    \centering
    \includegraphics[width=0.22\textwidth]{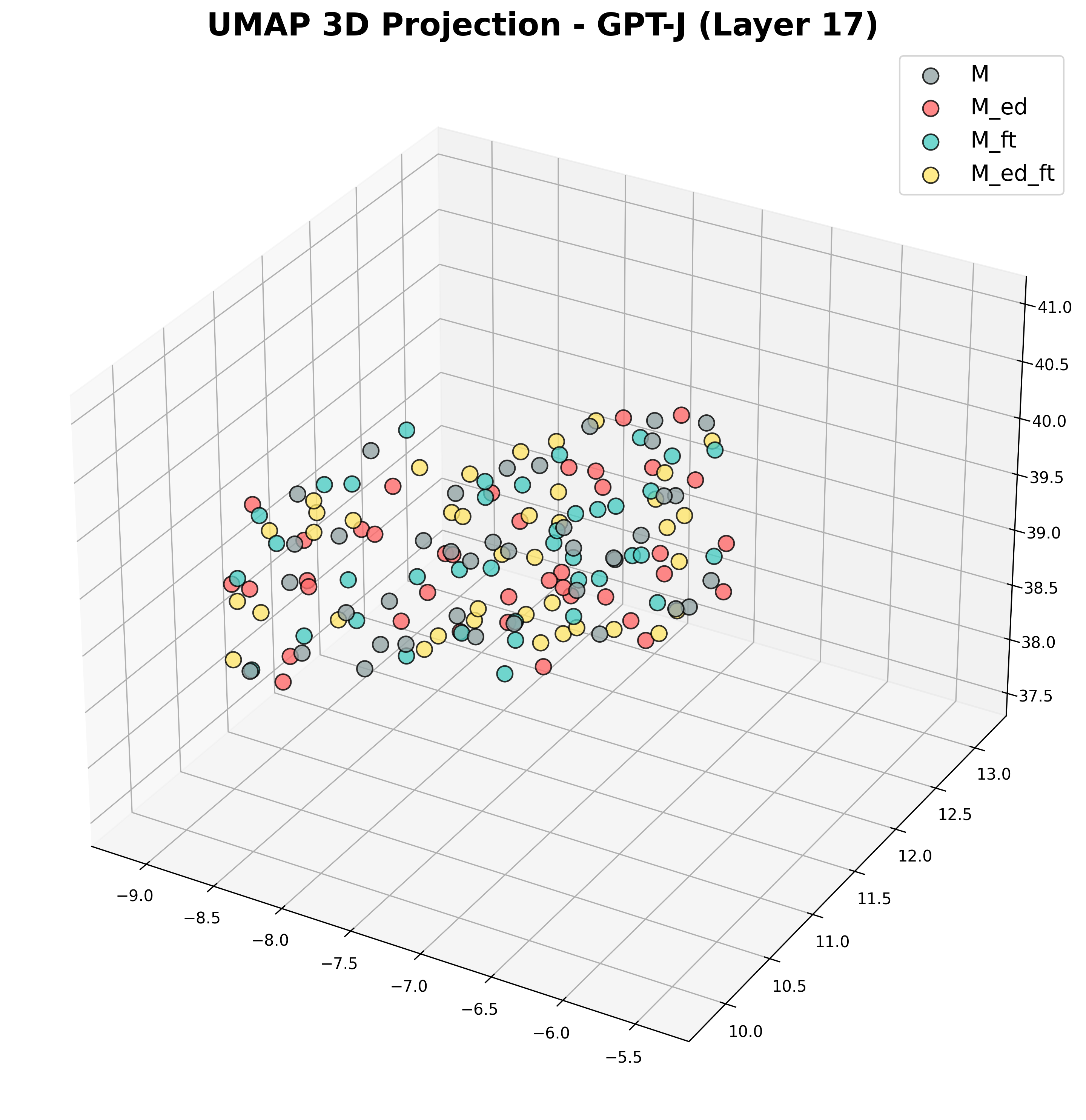}\hfill
    \includegraphics[width=0.22\textwidth]{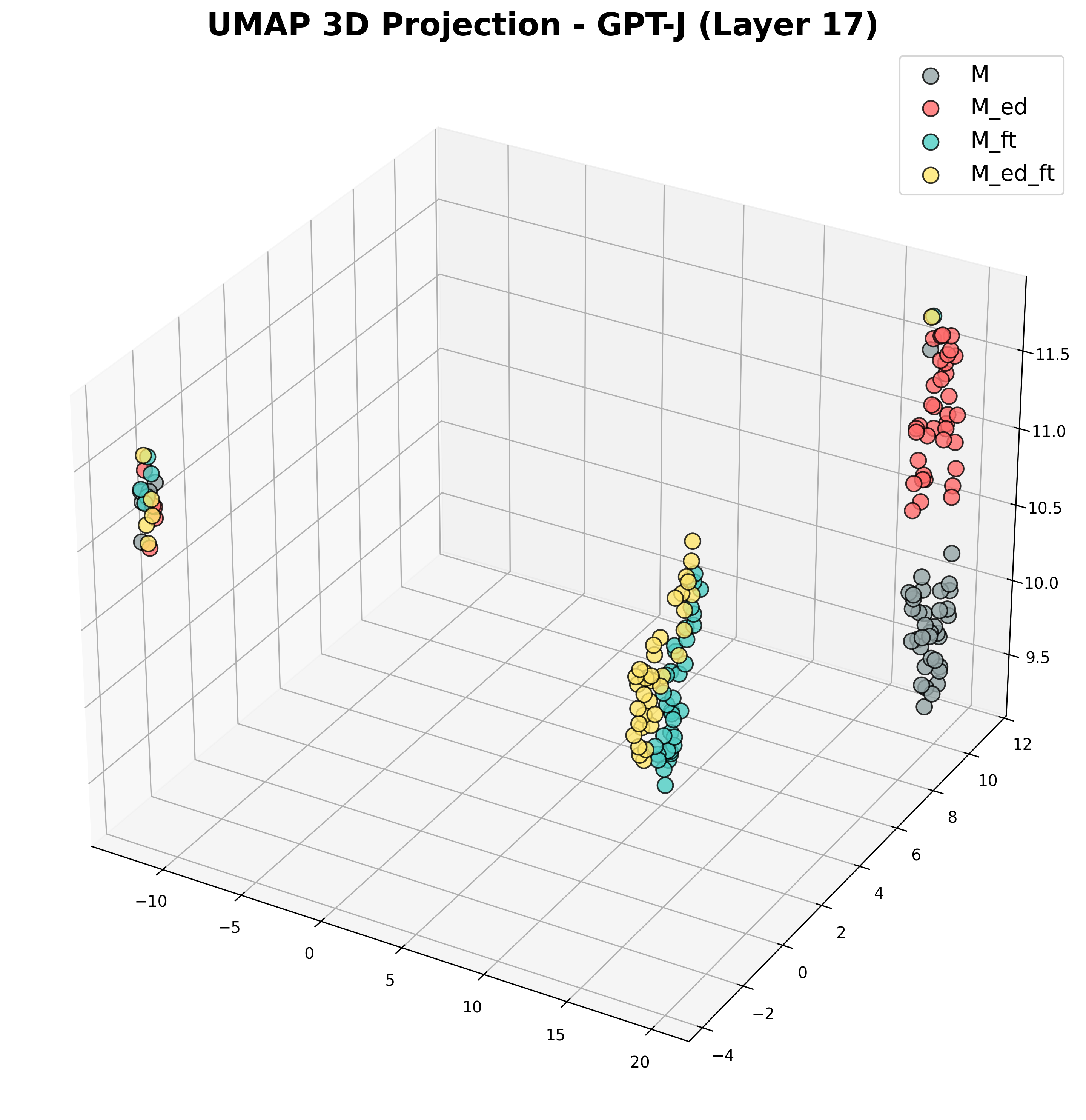}
    \caption{UMAP analysis of GPT-J using KE prompts (left) and FT prompts (right).}
    \label{fig:act_umap_single_source}
\end{figure}

Since the prompt set \(\mathcal{X}\) used consists two distinct sources, we further conduct analysis based on single-source prompts. The results of GPT-J is shown in Fig.~\ref{fig:act_umap_single_source} and the results of the remaining models are attached in App.~\ref{app:activation_res_breakdown}. As illustrated in Fig.~\ref{fig:act_umap_single_source}, activations corresponding to KE prompts are blended together across model categories (no clear grouping for either the KE or non‑KE models) while activations of FT prompts display a higher degree of separation between FT and non-FT models. This divergent pattern suggests that FT induces much larger changes in the model’s internal representations compared to KE.

Based on observed activation patterns of KE and FT, one possible mitigation for edit decay effect is to incorporate regularization that penalizes activation shifts specifically within the edited layers, thereby preventing the substantial FT‑induced changes from overwhelming the activation patterns introduced by KE~\citep{kirkpatrick2017overcoming}. Another potential approach leverages the concept of the null‑space~\citep{fang2025alphaeditnullspaceconstrainedknowledge}, projecting FT‑induced weight updates onto the orthogonal subspace of KE‑related parameters so that FT does not interfere with the edits.

\section{Conclusion}
In this paper, we show that the effect of knowledge edits often changes after fine-tuning. In many cases, fine-tuning degrades editing performance or even elicits new knowledge that differs from the target and the original knowledge. Different experimental configurations exhibit distinct degrees of edit decay, with smaller models and full-parameter fine-tuning generally experiencing more severe degradation of edits after fine-tuning. Our study also finds that a knowledge-rich FT dataset can further exacerbate KE performance degradation. In addition, we explored selective fine-tuning strategies for saving or removing edits and found that updating either only edited layers or non-edited layers can better remove edits. Moreover, we also observe that FT induces substantially larger activation shifts than KE, and we therefore propose two potential strategies to mitigate the resulting edit decay effect. Together, these findings provide empirical baselines and practical guidance for KE deployment. We suggest that future research on model editing should consider robustness across the entire LLM pipeline, rather than evaluating edits in isolation.

\clearpage
\bibliographystyle{ACM-Reference-Format}
\bibliography{sample-sigconf}

\appendix
\section{Knowledge Editing metrics}\label{app:KE_metrics_intro}
We adopt metrics from prior work \citep{meng2023locatingeditingfactualassociations, yao-etal-2023-editing}. For each \zsre edit $i$ in $M_\text{ed}$, let $s_i$, $r_i$, $o_i$ be the subject, relation, and target object, and $p(s_i, r_i)$ the base prompt. For \cf, $o_i^c$ denotes the original (real-world) object, and $\text{paraphrases}(s_i, r_i)$ and $\text{neighborhood}(s_i, r_i)$ the paraphrase and neighborhood prompt sets. Given prompt $p$, $Pr_\text{$M_\text{ed}$}(x \mid p)$ is the model's predicted \textit{probability} of token $x$. Although current KE metrics may be unreliable~\citep{liu2026evaluatingeditlocalityllm}, we focus on \textit{relative} changes before and after FT, reducing reliance on their absolute values.

\paragraph{ES Success (Efficacy / ES)}
Efficacy measures the proportion of successful edits. For \zsre, $i$ succeeds if $M_{ed}$ assigns the highest probability to $o_i$ under $p(s_i, r_i)$ (Equ.~\ref{eq:efficacy_def_zsre}). For \cf, $i$ succeeds if $M_{ed}$ assigns higher probability to $o_i$ than to $o_i^c$ under $p(s_i, r_i)$ (Equ.~\ref{eq:efficacy_def_mcf}):

{\tiny
\begin{equation}
\text{ES}_i^{\text{zsRE}} =
\mathbf{1}\Bigl(o_i = \arg\max_{x} Pr_\text{$M_\text{ed}$}\!\bigl(x \mid p(s_i, r_i)\bigr)
\Bigr)
\label{eq:efficacy_def_zsre}
\end{equation}
}

{\tiny
\begin{equation}
\text{ES}_i^{\text{CF}} =
\mathbf{1}\Bigl(
  Pr_\text{$M_\text{ed}$}\!\bigl(o_i \mid p(s_i, r_i)\bigr)
  >
  Pr_\text{$M_\text{ed}$}\!\bigl(o_i^c \mid p(s_i, r_i)\bigr)
\Bigr)
\label{eq:efficacy_def_mcf}
\end{equation}
}

Thus, the overall \textit{ES} can be calculated as:

{\tiny
\begin{equation}
\mathrm{ES}^\text{CF/zsRE}
= \frac{1}{N} \sum_{i=1}^N \text{ES}_i^{\text{CF/zsRE}}
\label{eq:efficacy_def_overall}
\end{equation}
}

\paragraph{Paraphrase Success (Paraphrase / PS)}
Paraphrase evaluates generalization after editing. For each instance $i$, we prepare a set of paraphrases of the base prompt $paraphrases(s_i, r_i)$. For \zsre, PS is the average top-1 accuracy over $paraphrases(s_i, r_i)$ (Equ.~\ref{eq:paraphrase_def_zsre}). For \cf, $p \in paraphrases(s_i, r_i)$ succeeds if the model prefers $o_i^c$ over the original $o_i$ (Equ.~\ref{eq:paraphrase_def_mcf}).

{\tiny
\begin{equation}
\mathrm{PS}^{\text{zsRE}}
= \frac{1}{N} \sum_{i=1}^N 
\mathbf{1}\!\left(
  o_i = \arg\max_{o} 
  Pr_\text{$M_\text{ed}$}\!\left( o \mid N(s_i, r_i) \right)
\right)
\label{eq:paraphrase_def_zsre}
\end{equation}
}

{\tiny
\begin{equation}
\text{PS}_i^{\text{CF}}
= \frac{1}{N} \sum_{i=1}^N
\mathbf{1}\Bigl(
  Pr_\text{$M_\text{ed}$}\!\bigl(o_i \mid p\bigr)
  >
  Pr_\text{$M_\text{ed}$}\!\bigl(o_i^c \mid p\bigr)
\Bigr)
\label{eq:paraphrase_def_mcf}
\end{equation}
}

\paragraph{Neighborhood Success (Specificity / NS)}
Specificity assesses edit locality by measuring unwanted impact on facts unrelated to the edit. For each instance $i$, we prepare a set of KE-irrelevant prompts $\text{neighborhood}(s_i, r_i)$. Equ.~\ref{eq:specificity_def_zsre} and~\ref{eq:specificity_def_mcf} give the NS definitions for \zsre and \cf, respectively, where $O(s_i, r_i)$ denotes the unrelated facts for \zsre:

{\tiny
\begin{equation}
\mathrm{NS}^{\text{zsRE}}
= \frac{1}{N} \sum_{i=1}^N 
\mathbf{1}\!\left(
  o_i = \arg\max_{o} 
  Pr_\text{$M_\text{ed}$}\!\left( o \mid O(s_i, r_i) \right)
\right)
\label{eq:specificity_def_zsre}
\end{equation}
}

{\tiny
\begin{equation}
\text{NS}^{\text{CF}} =
\frac{1}{N} \sum_{i=1}^N
\mathbf{1}\Bigl(
  Pr_\text{$M_\text{ed}$}\!\bigl(o_i \mid p\bigr)
  <
  Pr_\text{$M_\text{ed}$}\!\bigl(o_i^c \mid p\bigr)
\Bigr)
\label{eq:specificity_def_mcf}
\end{equation}
}

\section{Supplementary Results}\label{app:supplementary_results}

\subsection{KE results breakdown}\label{app:ke_perf_breakdown}
The full set of experimental combinations mentioned in below charts can be found in Tab.~\ref{tab:total_experiment_groups}. Detailed KE performances under FT of GPT2-XL, GPT-J, Llama2-7b and Llama3.1-8b are available at our Github repository (see \href{https://github.com/Cheng-Yinjie/edit_decay_KDD/tree/main/supplementary_materials/results_breakdown_ke}{KE results}).

\subsection{FT results breakdown}\label{app:ft_perf_breakdown}
Detailed downstream task performances of $M_{ed}$, $M_{ft}$ and $M_{ed\_ft}$ are presented in our \href{https://github.com/Cheng-Yinjie/edit_decay_KDD/tree/main/supplementary_materials/results_breakdown_ft}{repository} along with the extended observations.

\input{tabs/app/ds_perf_another_dst}
\subsection{Performance on HotpotQA}\label{app:ke_after_hotpotqa}
In this section, we evaluate the model's KE performance after fine-tuning on HotpotQA. The hyperparameters for this experiment are consistent with those used in the Commonsense dataset. We then compare the results with the performance of models fine-tuned on the Commonsense dataset. As shown in Tab.~\ref{tab:ds_perf_another}, the HotpotQA group consistently demonstrates lower KE performance compared to the Commonsense group. Regarding \textit{ES}, the largest performance gap is observed with GPT-J edited with 100 \zsre edits using MEMIT, with a difference of $14.31$. Based on result validation (see \href{https://github.com/Cheng-Yinjie/edit_decay_KDD/tree/main/supplementary_materials/results_validation}{Validation}), differences in performance are unlikely to stem from hyperparameter selections. It \textbf{could indicate that knowledge-rich datasets cause greater degradation in KE performance}. Further analysis is needed to explore this phenomenon.

\subsection{Performance on DeepSeek}\label{app:ke_about_deepseek}
When attempting to edit DeepSeek, we evaluated three configurations of editing layers: (1) layers identified via Causal Tracing with Frozen Components (CTFC) \citep{meng2023masseditingmemorytransformer}, (2) the layer settings used for LLaMA2, and (3) those used for GPT‑2 XL. 

The detailed results of layer selection, along with theoretical justification for these setups, are available in our \href{https://github.com/Cheng-Yinjie/edit_decay_KDD/tree/main/supplementary_materials/results_breakdown_deepseek}{repository}. The overall KE performance on the \zsre dataset, with edit counts ranging from $0$ to $10{,}000$, is also attached. For $10{,}000$ edits, all configurations perform poorly. Under the CTFC setting, increasing the number of edits from $1{,}000$ to $10{,}000$ causes a sharp decline in ES, from $20.17\%$ to just $0.4\%$. This degradation may arise from architectural differences between GPT‑series and Llama‑based models, as well as discrepancies in pre‑training data that hinder accurate simulation of the initial weights $W_0$~\citep{wang2025large}, ultimately reducing KE effectiveness. Moreover, once performance collapses after $10{,}000$ edits, it becomes highly vulnerable to further degradation during fine‑tuning, often approaching near‑zero accuracy. This \textbf{instability} prevents meaningful comparison of KE effectiveness before and after fine-tuning on the \zsre dataset. As a result, we do not perform additional fine-tuning experiments on DeepSeek.

\subsection{Only Fine-tuning Edited or Non-Edited Layers}\label{app:only_ft_selected_layers}
For FT with selected layers, we choose Llama2 and GPT-J for experiments. The detailed results along with its analysis are attached in our \href{https://github.com/Cheng-Yinjie/edit_decay_KDD/tree/main/supplementary_materials/results_breakdown_selective_layer_FT}{repository}.

\subsection{Activation related analysis}\label{app:activation_res_breakdown}
UMAP is a nonlinear dimensionality reduction method based on manifold theory that maps data into a lower dimensional space for visualization and clustering~\citep{mcinnes2020umapuniformmanifoldapproximation}. Eqn.~\ref{eq:umap_def} is the loss function used for dimensionality reduction, where $w_{h}$ is the weight of $e$ in the high dimensional case and $w_{l}$ is that in the low dimensional case for a single sample $e$ in all samples $E$.
Visualized analysis on activations of FT and KE prompts can be accessed \href{https://github.com/Cheng-Yinjie/edit_decay_KDD/tree/main/supplementary_materials/results_breakdown_activation_analysis}{here}.

{
\small 
\begin{gather}
\mathcal{L} = \sum_{e\in E} w_h(e) \log(\frac{w_h(e)}{w_l(e)})+(1-w_h(e))\log(\frac{1-w_h(e)}{1-w_l(e)})
\label{eq:umap_def} 
\end{gather}
}

\begin{figure}
    \centering
    \includegraphics[width=0.5\textwidth]{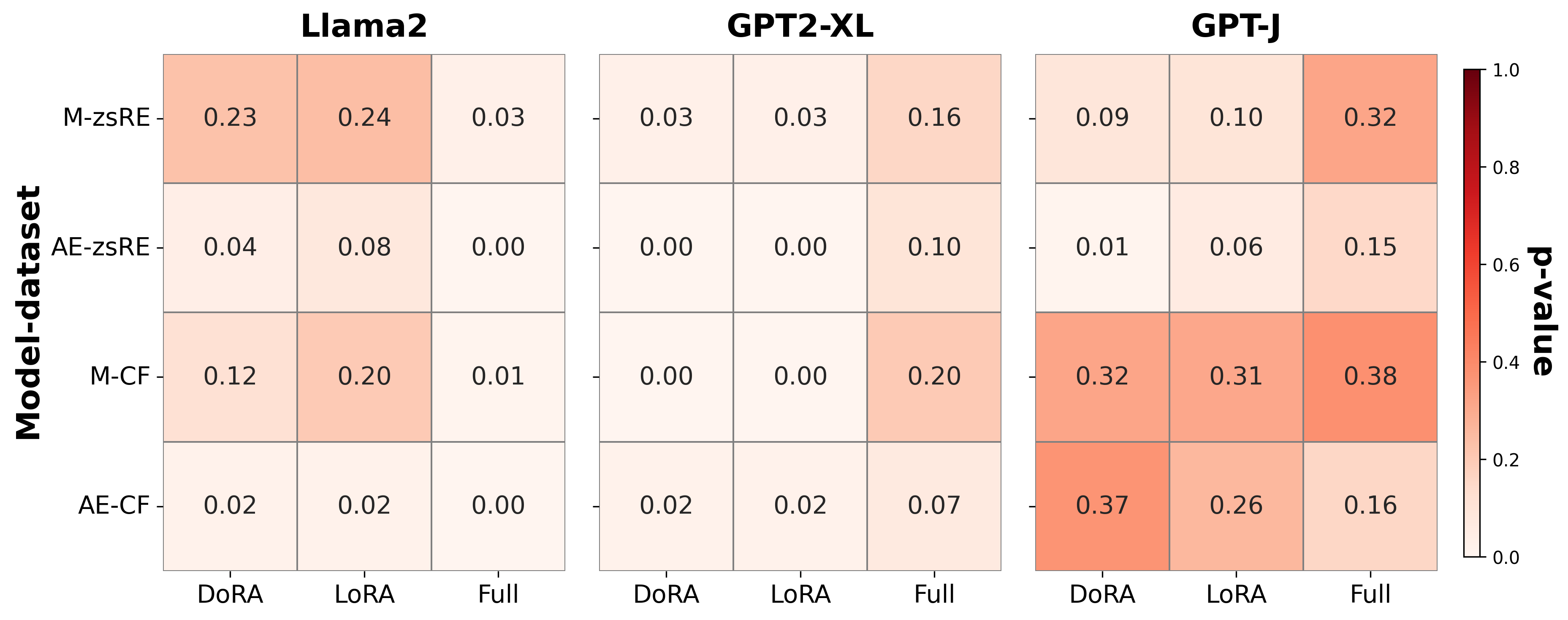}
    \caption{Significance test results across models. The format of Y label is \textit{KE method-KE dataset}, e.g. \aedit-\zsre means running AlphaEdit on \zsre dataset.}
    \label{fig:sign_test}
\end{figure}

\section{Statistical analysis}\label{sec:statistical_analysis}
\paragraph{Significance Test}
We apply a paired t‑test to assess the statistical significance of differences in KE performance between each $M_{ed}$ and its $M_{ed\_ft}$ counterpart. For model‑wise, we compute p‑values by models for each FT method using paired samples across edits from $100$ to $10,000$. As shown in Fig.~\ref{fig:sign_test}, GPT‑2 XL yields consistently low p‑values under both DoRA and LoRA, indicating that FT produces a genuine and substantial effect on KE performance. For Llama, full FT leads to the most pronounced performance reduction among all models. Most GPT‑J cases remain low with a few exceptions, suggesting that fine‑tuning exerts a strong and targeted influence on the model. For method‑wise, we calculate a p‑value by FT methods for each model. DoRA, LoRA, and full FT have p‑values of $3.29\times 10^{-5}$, $4.46\times 10^{-5}$, and $0.009$, respectively. These low values indicate that the performance differences are statistically significant and largely driven by the choice of FT method rather than variability KE setups.

\paragraph{Multi-seed run}
Since KE does not involve model re‑training, we apply multi‑seed runs only during the FT stage. For each PEFT method and each model, we add two additional seeds ($\{1, 2\}$), yielding 12 extra tests. Comparing the average downstream performance across seeds, the absolute differences are 3.7\%, 5.4\%, and 4.3\% for GPT‑2 XL, GPT‑J, and Llama2‑7B, respectively. Overall, the multi‑seed results show only modest variability across models, with performance remaining broadly stable.

\paragraph{Standard deviations}
\input{tabs/app/stat_analysis_std}
For KE, we compute the standard deviation of Efficacy for $M_{ed}$ across different sampled edits. For FT, we compute the standard deviation of average downstream performance across runs with different seeds. To save time, we adopt an edit number of $10,000$ to examine the deviation under the most critical circumstances. As shown in Tab.~\ref{tab:stat_analysis_std}, LoRA exhibits more stable performance than DoRA under varying seeds, while for KE methods, AlphaEdit shows a smaller standard deviation than MEMIT.

\section{Reproducibility}\label{sec:reproducibility}
Environment specification and implementation examples can be accessed in our \href{https://github.com/Cheng-Yinjie/edit_decay_KDD}{repository}.

\end{document}

%% file: tabs/app/total_experiment_groups.tex
\begin{table}[t]
\centering
\tiny
\caption{Scope of the experimental parameters. M - models, D - datasets, E - editing methods, N - editing numbers, F - fine-tuning methods.}

\begin{tabular}{|c|c|c|c|c|c|c|c|c|}
\noalign{\hrule}
\textbf{Model} & \multirow{6}{*}{\textbf{$\times$}} & \textbf{Dataset} & \multirow{6}{*}{\textbf{$\times$}} & \textbf{Edit method} & \multirow{6}{*}{\textbf{$\times$}} & \textbf{Edits} & \multirow{6}{*}{\textbf{$\times$}} & \textbf{fine-tune method} \\
\cline{1-1} \cline{3-3} \cline{5-5} \cline{7-7} \cline{9-9}
GPT2-XL &  &  &  & No editing &  & 0 &  & No fine-tuning \\
GPTJ &  & zsRE &  & MEND &  & 100 &  & LoRA \\
Llama2 &  & COUNTERFACT &  & MEMIT &  & 1000 &  & DoRA \\
Llama3.1 &  &  &  & AlphaEdit &  & 10000 &  & Full-size \\
DeepSeek &  &  &  &  &  &  &  &  \\
\noalign{\hrule}
\end{tabular}

\vspace{0.3em}

{\tiny
\begin{align*}
\begin{aligned}
S &= M \times D \times E \times N \times F \\
&= \{(m, d, e, n, f) \mid m \in M, d \in D, e \in E, n \in N, f \in F\}
\end{aligned}
\end{align*}
}

\label{tab:total_experiment_groups}
\end{table}

%% file: tabs/edit_perf_ft_wise.tex
\begin{table}[b]
\setlength{\tabcolsep}{1.1pt}
\centering
\scriptsize
\renewcommand{\arraystretch}{0.5}
\caption{Success rate (ES, \%) of edited GPT-J, GPT2-XL and Llama2  without finetuning (No ft) and with fine-tuning (LoRA, DoRA, Full ft). Full results across KE metrics are in App.~\ref{app:ke_perf_breakdown}.}
\begin{tabular}{lrcccccccccccc}
\toprule
 &  & \multicolumn{4}{c}{\textbf{GPT-J}} & \multicolumn{4}{c}{\textbf{GPT2-XL}} & \multicolumn{4}{c}{\textbf{Llama2}} \\
\cmidrule(lr){3-6}\cmidrule(lr){7-10}\cmidrule(lr){11-14}
 & \#Edits & No ft & LoRA & DoRA & Full ft & No ft & LoRA & DoRA & Full ft & No ft & LoRA & DoRA & Full ft \\
\midrule
\multirow{3}{*}{\begin{tabular}{@{}c@{}}\memit \\on \\\zsre\end{tabular}} & $10^2$ & 99.07 & \textcolor{darkgreen}{88.79} & \textcolor{darkgreen}{89.87} & \textcolor{darkgreen}{97.95} & 80.00 & \textcolor{darkgreen}{58.37} & \textcolor{darkgreen}{58.39} & \textcolor{darkpink}{81.16} & 86.03 & \textcolor{darkgreen}{76.30} & \textcolor{darkgreen}{72.00} & \textcolor{darkgreen}{22.67}\\
 & $10^3$ & 99.10 & \textcolor{darkgreen}{84.53} & \textcolor{darkgreen}{85.31} & \textcolor{darkgreen}{98.74} & 77.85 & \textcolor{darkgreen}{43.51} & \textcolor{darkgreen}{45.18} & \textcolor{darkpink}{81.22} & 51.38 & \textcolor{darkgreen}{46.52} & \textcolor{darkgreen}{46.00} & \textcolor{darkgreen}{10.22} \\
 & $10^4$ & 96.63 & \textcolor{darkgreen}{66.52} & \textcolor{darkgreen}{67.83} & \textcolor{darkgreen}{89.38} & 62.61 & \textcolor{darkgreen}{20.34} & \textcolor{darkgreen}{20.65} & \textcolor{darkpink}{63.39} & 48.62 & \textcolor{darkpink}{48.64} & \textcolor{darkgreen}{48.23} & \textcolor{darkgreen}{14.00} \\
\midrule
\multirow{3}{*}{\begin{tabular}{@{}c@{}}\aedit \\on \\\zsre\end{tabular}} & $10^2$ & 99.33 & \textcolor{darkgreen}{84.74} & \textcolor{darkgreen}{99.33} & \textcolor{darkgreen}{98.50} & 97.18 & \textcolor{darkgreen}{67.80} & \textcolor{darkgreen}{76.72} & \textcolor{darkgreen}{18.04} & 93.33 & \textcolor{darkgreen}{57.96} & \textcolor{darkgreen}{93.33} & \textcolor{darkgreen}{21.83} \\
 & $10^3$ & 99.31 & \textcolor{darkgreen}{84.74} & \textcolor{darkgreen}{81.70} & \textcolor{darkgreen}{98.98} & 93.13 & \textcolor{darkgreen}{54.52} & \textcolor{darkgreen}{55.11} & \textcolor{darkgreen}{24.74} & 93.23 & \textcolor{darkgreen}{50.45} & \textcolor{darkgreen}{51.53} & \textcolor{darkgreen}{24.36} \\
 & $10^4$ & 89.81 & \textcolor{darkgreen}{49.97} & \textcolor{darkgreen}{23.64} & \textcolor{darkgreen}{74.62} & 62.34 & \textcolor{darkgreen}{25.80} & \textcolor{darkgreen}{27.47} & \textcolor{darkgreen}{22.09} & 84.31 & \textcolor{darkgreen}{46.03} & \textcolor{darkgreen}{45.38} & \textcolor{darkgreen}{24.44} \\
\midrule
\midrule
\multirow{3}{*}{\begin{tabular}{@{}c@{}}\memit \\on \\\cf\end{tabular}} & $10^2$ & 100.00 & \textcolor{darkgreen}{100.00} & \textcolor{darkgreen}{100.00} & \textcolor{darkgreen}{100.00} & 97.00 & \textcolor{darkgreen}{83.00} & \textcolor{darkgreen}{82.00} & \textcolor{darkgreen}{97.00} & 100.00 & \textcolor{darkgreen}{94.00} & \textcolor{darkgreen}{97.00} & \textcolor{darkgreen}{47.00}\\
 & $10^3$ & 100.00 & \textcolor{darkgreen}{99.00} & \textcolor{darkgreen}{99.00} & \textcolor{darkgreen}{99.00} & 93.40 & \textcolor{darkgreen}{78.37} & \textcolor{darkgreen}{78.50} & \textcolor{darkgreen}{92.60} & 51.38 & \textcolor{darkgreen}{46.52} & \textcolor{darkgreen}{46.00} & \textcolor{darkgreen}{10.22} \\
 & $10^4$ & 99.10 & \textcolor{darkgreen}{94.34} & \textcolor{darkgreen}{94.46} & \textcolor{darkgreen}{97.79} & 79.17 & \textcolor{darkgreen}{62.10} & \textcolor{darkgreen}{61.97} & \textcolor{darkgreen}{78.03} & 86.96 & \textcolor{darkgreen}{70.18} & \textcolor{darkgreen}{68.27} & \textcolor{darkgreen}{48.07} \\
\midrule
\multirow{3}{*}{\begin{tabular}{@{}c@{}}\aedit \\on \\\cf\end{tabular}} & $10^2$ & 100.00 & \textcolor{darkgreen}{99.00} & \textcolor{darkgreen}{100.00} & \textcolor{darkgreen}{100.00} & 100.00 & \textcolor{darkgreen}{96.00} & \textcolor{darkgreen}{98.00} & \textcolor{darkgreen}{19.00} & 100.00 & \textcolor{darkgreen}{66.00} & \textcolor{darkgreen}{61.00} & \textcolor{darkgreen}{48.00} \\
 & $10^3$ & 98.35 & \textcolor{darkpink}{99.70} & \textcolor{darkpink}{99.60} & \textcolor{darkgreen}{97.95} & 100.00 & \textcolor{darkgreen}{89.30} & \textcolor{darkgreen}{90.55} & \textcolor{darkgreen}{21.92} & 99.10 & \textcolor{darkgreen}{52.40} & \textcolor{darkgreen}{56.90} & \textcolor{darkgreen}{48.60} \\
 & $10^4$ & 98.87 & \textcolor{darkpink}{99.16} & \textcolor{darkgreen}{82.00} & \textcolor{darkgreen}{95.85} & 92.94 & \textcolor{darkgreen}{53.91} & \textcolor{darkgreen}{57.65} & \textcolor{darkgreen}{21.92} & 87.43 & \textcolor{darkgreen}{37.90} & \textcolor{darkgreen}{37.15} & \textcolor{darkgreen}{48.46} \\
\bottomrule
\end{tabular}
\label{tab:edit_perf_ft_wise}
\end{table}

%% file: tabs/app/over_ke_perf_mend.tex
\begin{table}[t]
\centering
\small
\caption{KE performance (\%) using MEND under different FT settings (No fine-tuning and DoRA) across datasets (\zsre and \cf) and models}
\begin{tabular}{ccllll}
\toprule
\multirow{2}{*}{Model} & \multirow{2}{*}{\#Edits} & \multicolumn{2}{c}{zsRE} & \multicolumn{2}{c}{COUNTERFACT} \\ 
\cmidrule(lr){3-4}
\cmidrule(lr){5-6}
 &  & No ft & DoRA & No ft & DoRA \\ \cmidrule(lr){1-6}
\multirow{2}{*}{GPT2-XL} & $10^3$ & 66.57 & 48.92 & 0.00 & 0.00 \\
 & $10^4$ & 52.70 & 34.88 & 0.00 & 0.00 \\ \cmidrule(lr){1-6}
\multirow{2}{*}{GPT-J} & $10^3$ & 68.32 & 49.77 & 0.33 & 0.00 \\ 
 & $10^4$ & 45.27 & 27.55 & 0.64 & 0.00 \\ \cmidrule(lr){1-6}
\multirow{2}{*}{Llama2} & $10^3$ & 71.15 & 53.02 & 0.52 & 0.00 \\ 
 & $10^4$ & 53.24 & 35.88 & 0.31 & 0.00 \\ \cmidrule(lr){1-6}
\multirow{2}{*}{Llama3.1} & $10^3$ & 82.15 & 63.44 & 0.73 & 0.00 \\
 & $10^4$ & 62.17 & 44.55 & 0.57 & 0.00 \\
\bottomrule
\end{tabular}
\label{tab:over_ke_perf_mend}
\end{table}

%% file: tabs/decrease_rate_after_ft.tex
\begin{table}[t]
\centering
\scriptsize
\setlength{\tabcolsep}{1.5pt}
\caption{Decreasing rate in KE performance (($ES_{\text{ed}}$ - $ES_{\text{ed\_ft}}$) / $ES_{\text{ed}}$, \%) after FT (\zsre). \memit, \aedit refer to MEMIT and AlphaEdit. We report average per model per FT method (Avg.), and average across models.}
\resizebox{0.45\textwidth}{!}{
\begin{tabular}{lccccccccc}
\toprule
 & \multicolumn{3}{c}{\textbf{GPT-J}} & \multicolumn{3}{c}{\textbf{GPT2-XL}} & \multicolumn{3}{c}{\textbf{LLAMA2}} \\ \midrule
KE-\#Edits & DoRA & LoRA & Full  & DoRA  & LoRA & Full & DoRA & LoRA & Full  \\ \midrule
\memit-$10^2$ & 10.38 & 9.29 & 1.13  & 27.04  & 27.01 & -1.45 & 11.31 & 16.31 & 73.65  \\
\memit-$10^3$  & 14.70 & 13.92 & 0.36  & 44.11  & 41.97 & -4.33 & 9.46 & 10.47 & 80.11  \\
\memit-$10^4$ & 31.16 & 29.80 & 7.50  & 67.51  & 67.02 & -1.25 & 3.66 & 0.80 & 71.21  \\ \midrule
\aedit-$10^2$  & 14.69 & 0.00 & 0.84  & 30.23  & 21.05 & 81.44 & 37.90 & 0.00 & 76.61 \\
\aedit-$10^3$& -0.02 & 17.73 & 16.91  & 41.46  & 40.82 & 73.43 & 45.89 & 44.73 & 73.87  \\
\aedit-$10^4$ & 44.36 & 73.68 & 16.91  & 58.61  & 55.94 & 64.57 & 45.40 & 46.17 & 71.01  \\ \midrule
Avg. & 19.21 & 24.07 & 4.51 & 44.83  & 42.30 & 35.40  & 25.60 & 19.75 & 74.41 \\ \midrule
Overall Avg.  & \multicolumn{3}{c}{15.93 {\tiny $\pm$ 19.04}}  & \multicolumn{3}{c}{40.84 {\tiny $\pm$ 26.24}}  & \multicolumn{3}{c}{39.92 {\tiny $\pm$ 29.64}} \\ \bottomrule
\end{tabular}}
\label{tab:decrease_rate_after_ft}
\end{table}

%% file: tabs/efr_gptj.tex
\begin{table}[t]
\centering
\footnotesize
\setlength{\tabcolsep}{3.05pt}
\renewcommand{\arraystretch}{0.79}
\caption{Edit Flip Ratio (EFR, \%) for GPT-J across FT methods. Decrement $\Delta$ES is equal to $M_{\text{ed}}$'s ES minus $M_{\text{ed\_ft}}$'s ES, raw ES results are in App.~\ref{app:ke_perf_breakdown}. Higher EFR values indicate more removal of original success edits.}
\begin{tabular}{ccccccccc}
\toprule
\multirow{2}{*}{Dataset} & \multirow{2}{*}{KE} & \multirow{2}{*}{\#Edits} & \multicolumn{2}{c}{LoRA} & \multicolumn{2}{c}{DoRA} & \multicolumn{2}{c}{Full ft} \\
 &  &  & $\Delta$ES & EFR & $\Delta$ES & EFR & $\Delta$ES & EFR \\ 
 \midrule
\multirow{6}{*}{\zsre} & \multirow{3}{*}{\memit} & $10^2$ & 10.28 & 5.00 & 9.20 & 5.00 & 1.12 & 0.00 \\
 &  & $10^3$ & 14.57 & 5.51 & 13.79 & 5.71 & 0.36 & 0.1 \\
 &  & $10^4$ & 30.41 & 15.60 & 29.10 & 14.84 & 7.55 & 3.62 \\
 \cmidrule{2-9} 
 & \multirow{3}{*}{\aedit} & $10^2$ & 14.59 & 0.00 & 0.00 & 0.00 & 0.83 & 0.00 \\
 &  & $10^3$ & 14.57 & 5.81 & 17.61 & 7.01 & 0.33 & 0.00 \\
 &  & $10^4$ & 39.84 & 25.27 & 66.17 & 0.00 & 15.19 & 9.92 \\ 
 \midrule
\multirow{6}{*}{\cf} & \multirow{3}{*}{\memit} & $10^2$ & 0.00 & 3.00 & 0.00 & 2.00 & 0.00 & 0.00 \\
 &  & $10^3$ & 1.00 & 8.12 & 1.00 & 6.82 & 0.10 & 0.20 \\
 &  & $10^4$ & 4.76 & 22.39 & 4.64 & 20.95 & 1.31 & 6.03 \\
 \cmidrule{2-9} 
 & \multirow{3}{*}{\aedit} & $10^2$ & 1.00 & 2.00 & 0.00 & 2.00 & 0.00 & 0.00 \\
 &  & $10^3$ & -1.35 & 4.31 & -1.25 & 4.11 & 0.40 & 0.00 \\
 &  & $10^4$ & -0.29 & 3.49 & 16.87 & 0.46 & 3.02 & 18.18 \\ 
 \bottomrule
\end{tabular}
\label{tab:efr_gptj}
\end{table}

%% file: tabs/quality_analysis_examples.tex
\begin{table*}
\centering
\small
\fontsize{8}{9}\selectfont
\setlength{\tabcolsep}{1pt}
\renewcommand{\arraystretch}{1.0}
\caption{Examples of model behaviors on the editing target knowledge before and after fine-tuning. \textbf{Erased edits} refer to the cases where a success edit is erased after fine-tuning. \textbf{Stable edits} refer to target edits that are successfully introduced and retained after fine-tuning. \textbf{Emergent Edits} are cases where the target edits initially fail but emerge after fine-tuning. \textbf{Impossible edits} are those where the target knowledge is never successfully introduced, either immediately after editing or following further fine-tuning.}
\begin{tabular}{p{2cm}p{5cm}p{0.1cm}p{1.6cm}p{0.1cm}p{1.5cm}p{0.1cm}p{1.5cm}p{0.1cm}p{2cm}p{0.1cm}cp{0.1cm}}
\toprule
\textbf{Category} & \textbf{Prompt Context} & & \textbf{$M$ output} & &  \textbf{Target} & &  \textbf{$M_{\text{ed}}$ output} & &  \textbf{$M_{\text{ed\_ft}}$ output} & &  \textbf{Data} & \\
\midrule
\multirow{3}{*}{\centering\arraybackslash\textbf{Stable Edits}} & Mother tongue of Danielle Darrieux is &  & French & &  English & &  English & &  English & &  \cf &  \\
& Official religion of Edwin of Northumbria is &  & Christianity & &  Islam & &  Islam & &  Islam & &  \cf &  \\
& Toko Yasuda, the &  & guitar & &  piano & &  piano & &  piano & &  \zsre &  \\
\midrule

\multirow{3}{*}{\centering\arraybackslash\textbf{Erased Edits}} & Which family does Ramalinaceae belong to? &  & Lamiales & &  Lecanorales & &  Lecanorales & &  Ramalinaceae & &  \zsre &  \\
& Savdhaan India @ 11, formulated in &  & India & &  Poland & &  Poland & &  India & &  \cf &  \\
& Laurent Cars was employed in &  & Paris & &  Philadelphia & &  Philadelphia & &  London & &  \cf &  \\
\midrule

\multirow{2}{*}{\centering\arraybackslash\textbf{Emergent Edits}} & Mother tongue of Danielle Darrieux is &  & French & &  English & &  United States & &  English & &  \cf &  \\
& Native language of Symeon of Polotsk is &  & Russian & &  French & &  Russian & &  French & &  \cf &  \\
\midrule
\multirow{3}{*}{\centering\arraybackslash\textbf{Impossible Edits}} & In which state is Qaleh Lan located? &  & Kermanshah, Iran & &  Poshtdarband RD & &  Qaleh Zari County & &  Qaleh Zari & &  \zsre &  \\
& Date of birth of Priyankara Wickramasinghe? &  & Priyankara W. & &  12 May 1977 & &  1 May 1977 & &  1 May 1977 & &  \zsre & \\
& The voice type of Gemma Bosini is what? &  & singer & &  soprano & &  Au-natural & &  Au-natural & &  \zsre & \\
\bottomrule
\end{tabular}
\label{tab:quality_analysis_examples}
\end{table*}

%% file: tabs/ft_selected_layers_keperf.tex
\begin{table*}[t]
\centering
\small
\caption{KE performance (\%) of Llama2 being edited using \aedit on \cf, and then being fine-tuned with selective layers. $M$\textsuperscript{1} for model without editing or fine-tuning; $M_\textit{ed}$\textsuperscript{1} for edited-only model; $M_\textit{ed\_ft\_all}$\textsuperscript{3} for edited-then-finetuned with all layers; $M_\textit{ed\_ft\_edited}$\textsuperscript{4} for edited-then-finetuned with edited layers; $M_\textit{ed\_ft\_non-edited}$\textsuperscript{5} for edited-then-finetuned with non-edited layers. ES, NS and PS are KE metrics, DS is the average score of downstream tasks.}
\adjustbox{width=0.99\textwidth,center}
{
\begin{tabular}{cclcccclcccc}
\toprule
\multirow{2}{*}{\begin{tabular}[c]{@{}c@{}}KE \\ performance\end{tabular}} & Llama2 &  & \multicolumn{4}{c}{100 Edits} &  & \multicolumn{4}{c}{1000 Edits} \\
 & $M$\textsuperscript{1} &  & $M_\text{ed}$\textsuperscript{2} & $M_\textit{ed\_ft\_all}$\textsuperscript{3} & $M_\textit{ed\_ft\_edited}$\textsuperscript{4} & $M_\textit{ed\_ft\_non-edited}$\textsuperscript{5} &  & $M_\text{ed}$ & $M_\textit{ed\_ft\_all}$ & $M_\textit{ed\_ft\_edited}$ & $M_\textit{ed\_ft\_non-edited}$\\
\midrule
ES & 20.00 &  & 96.00 & 98.00 & 66.00 & 72.00 &  & 100.00 & 90.55 & 57.80 & 60.30 \\
PS & 35.00 &  & 87.50 & 93.00 & 68.00 & 64.00 &  & 95.75 & 77.03 & 60.95 & 54.40 \\
NS & 69.00 &  & 76.60 & 76.60 & 83.60 & 82.20 &  & 72.44 & 73.20 & 79.52 & 80.31 \\
DS & 1.77 & & 2.15 & 81.7 & 65.43 & 80.61 & & 4.79 & 81.00 & 65.35 & 72.46 \\
\bottomrule
\end{tabular}
}
\label{tab:ft_selected_layers_keperf}
\end{table*}

%% file: tabs/decrease_rate_caused_by_KE.tex
\begin{table}[t]
\centering
\scriptsize
\setlength{\tabcolsep}{1.2pt}
\caption{Average and Standard deviation of degradation (\%) in evaluation score described in Sec.~\ref{sec:eval_methods} across models and fine-tuning methods. Avg.\textsuperscript{1}, Std.\textsuperscript{2} are metrics for individual model; Avg\_m\textsuperscript{3}, Std\_m\textsuperscript{4} are across models.}
\begin{tabular}{cccccc}
\toprule
\textbf{Model} & \textbf{Metrics} & \textbf{No ft} & \textbf{Full ft} & \textbf{LoRA} & \textbf{DoRA} \\
\midrule
\multirow{2}{*}{GPT-J} & Avg.\textsuperscript{1} & 56.84 & 7.17 & 5.19 & 5.87 \\
 & Std.\textsuperscript{2} & 17.82 & 12.72 & 8.68 & 4.66 \\ 
\midrule
\multirow{2}{*}{Llama2} & Avg. & -206.92 & 37.27 & 10.81 & 5.93 \\
 & Std. & 220.54 & 8.05 & 15.17 & 8.00 \\
\midrule
\multirow{2}{*}{GPT2-XL} & Avg. & 1.70 & -0.38 & 2.07 & 8.85 \\
 & Std. & 23.66 & 2.87 & 9.05 & 9.32 \\
\midrule
\multicolumn{2}{c}{Avg\_m\textsuperscript{3}} & -49.46 & 14.69 & 6.02 & 6.88 \\
\multicolumn{2}{c}{Std\_m\textsuperscript{4}} & 169.81 & 18.6 & 11.63 & 7.50 \\
\bottomrule
\end{tabular}
\label{tab:decrease_rate_caused_by_KE}
\end{table}

%% file: tabs/app/ds_perf_another_dst.tex

\begin{table}[b]
\setlength{\tabcolsep}{1.1pt}
\centering
\scriptsize
\renewcommand{\arraystretch}{0.4}
\caption{KE performance (\%) after fine-tuning using Commonsense\textsuperscript{1} and HotpotQA\textsuperscript{2} datasets.}
\begin{tabular}{ccccccccccccccc}
\toprule
\#Edits & KE & Model   & Dataset & FT & ES (HQA) & ES (CS\textsuperscript{1}) & PS (HQA\textsuperscript{2}) & PS (CS) & NS (HQA) & NS (CS) \\
\midrule
$10^{2}$ & \memit     & Llama2  & zsRE    & DoRA      & 72.76    & 76.3        & 69.65    & 71.69       & 33.02    & 29.69       \\
$10^{4}$ & \aedit & Llama2  & CF      & LoRA      & 68       & 70.18       & 48.4     & 64.32       & 83.67    & 62.35       \\
$10^{2}$ & \memit     & GPT-J   & zsRE    & DoRA      & 75.56    & 89.87       & 70.63    & 84.62       & 35.71    & 27.64       \\
$10^{3}$ & \memit & GPT-J   & CF      & LoRA      & 100      & 99.7        & 87.7     & 90.9        & 82.6     & 79.58       \\
$10^{3}$ & \memit     & GPT2-XL & zsRE    & DoRA      & 39.84    & 45.18       & 27.83    & 43.62       & 29.15    & 24.35       \\
$10^{4}$ & \aedit & GPT2-XL & CF      & LoRA      & 41.56    & 53.91       & 32.19    & 41.94       & 78.15    & 71.81       \\
\bottomrule
\end{tabular}
\label{tab:ds_perf_another}
\end{table}

%% file: tabs/app/stat_analysis_std.tex
\begin{table}[t]
\centering
\scriptsize
\caption{Standard deviation analysis: Avg. DS Perf = average downstream performance; SD = standard deviation.}

\begin{tabular}{c|cc|c}
Model     & Metric       & Method  & SD (\%)   \\ 
\toprule
GPT2-XL   & Efficacy     & \memit-zsRE  & 0.33 \\
          &              & \memit-CF    & 0.22 \\
          &              & \aedit-zsRE & 0.33 \\
          &              & \aedit-CF   & 0.24 \\
          & Avg. DS Perf & DoRA    & 8.93 \\
          &              & LoRA    & 5.27 \\ 
\midrule
GPT-J     & Efficacy     & \memit-zsRE  & 0.16 \\
          &              & \memit-CF    & 0.11 \\
          &              & \aedit-zsRE & 0.14 \\
          &              & \aedit-CF   & 0.09 \\
          & Avg. DS Perf & DoRA    & 4.83 \\
          &              & LoRA    & 3.71 \\ 
\hline
LLama2-7b & Efficacy     & \memit-zsRE  & 0.36 \\
          &              & \memit-CF    & 0.49 \\
          &              & \aedit-zsRE & 0.15 \\
          &              & \aedit-CF   & 0.13 \\
          & Avg. DS Perf & DoRA    & 2.13 \\
          &              & LoRA    & 1.77 \\ 
\bottomrule
\end{tabular}

\label{tab:stat_analysis_std}
\end{table}